%% file: main.tex
\documentclass[10pt,twocolumn,letterpaper]{article}

\usepackage{cvpr}              
\usepackage{booktabs} 
\usepackage{makecell} 
\usepackage{amsfonts,amssymb} 
\usepackage{amsmath}
\usepackage{mathtools}
\usepackage{comment}
\usepackage{bbm}
\usepackage{bm}
\usepackage{multicol}
\usepackage{threeparttable}
\DeclareMathAlphabet{\mathsfit}{\encodingdefault}{\sfdefault}{m}{sl}
\SetMathAlphabet{\mathsfit}{bold}{\encodingdefault}{\sfdefault}{bx}{n}

\usepackage{color}
\usepackage[table]{xcolor}
\definecolor{mygray}{gray}{0.95}
\usepackage{caption}
\usepackage{subcaption}
\usepackage{multirow}
\usepackage{ulem}
\usepackage{cuted}
\usepackage{tabularx}
\usepackage{fontawesome5}
\newcommand \footnoteONLYtext[1]
{
	\let \mybackup \thefootnote
	\let \thefootnote \relax
	\footnotetext{#1}
	\let \thefootnote \mybackup
	\let \mybackup \imareallyundefinedcommand
} 

\usepackage[pagebackref,breaklinks=true,colorlinks,citecolor=blue,urlcolor=blue,linkcolor=blue,bookmarks=false]{hyperref}

\input{preamble}
\definecolor{cvprblue}{rgb}{0.21,0.49,0.74}

\def\paperID{} 
\def\confName{CVPR}
\def\confYear{2026}

\title{Superman: Unifying Skeleton and Vision\\for Human Motion Perception and Generation
}

\author{
Xinshun Wang$^{1,*}$, Peiming Li$^{1,*}$, Ziyi Wang$^{1,*}$,\\Zhongbin Fang$^{1}$, Zhichao Deng$^{1}$, Songtao Wu$^{2}$, Jason Li$^{3}$, Mengyuan Liu$^{1}$ \\
    \small{$^{1}$State Key Laboratory of General Artificial Intelligence, Peking University, Shenzhen Graduate School}\\
    \small{$^{2}$Sony R\&D Center}
    \small{$^{3}$Nanyang Technological University} \\
    \small{\url{https://github.com/BradleyWang0416/Superman}}
    \vspace{-2.5mm}
}

\begin{document}

\twocolumn[{%
\renewcommand\twocolumn[1][]{#1}%
\maketitle
\centering
\captionsetup{type=figure}
\vspace{-6mm}
\includegraphics[width=0.99\textwidth]{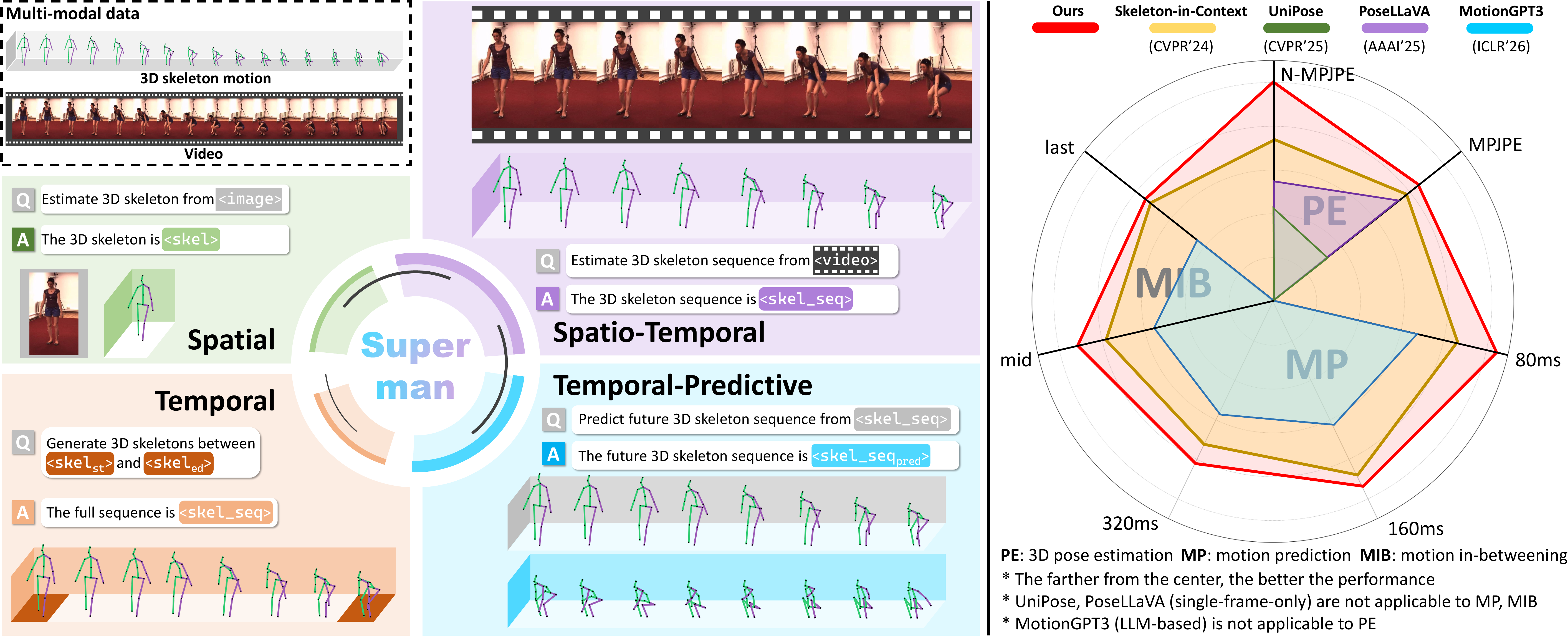}
\vspace{-1.5mm}
\caption{\textbf{A Unified Framework for Human Motion Perception and Generation.} Our unified model achieves state-of-the-art performance across three traditionally disparate tasks. Our method consistently outperforms existing methods across all tasks simultaneously: \textbf{(1) 3D Pose Estimation}, \textbf{(2) Motion Prediction}, and \textbf{(3) Motion In-betweening}.}
\label{fig:teaser}
\vspace{5mm}
}]



\input{sec_cameraready/0_abstract}
\input{sec_cameraready/1_intro}

\input{sec_cameraready/2_related_work}

\input{sec_cameraready/3_method}
\input{sec_cameraready/4_exp}
\input{sec_cameraready/5_conclusion}



\normalem
{
    \small
    \bibliographystyle{ieeenat_fullname}
    \bibliography{main}
}

\input{sec_cameraready/X_suppl}

\end{document}

%% file: sec_cameraready/0_abstract.tex
\begin{abstract}
\footnoteONLYtext{
*~Equal contribution. Corresponding author is Mengyuan Liu.}
Human motion analysis tasks, such as temporal 3D pose estimation, motion prediction, and motion in-betweening, play an essential role in computer vision.
However, current paradigms suffer from severe fragmentation.
First, the field is split between ``perception'' models that understand motion from video but only output text, and ``generation'' models that cannot perceive from raw visual input.
Second, generative MLLMs are often limited to single-frame, static poses using dense, parametric SMPL models, failing to handle temporal motion.
Third, existing motion vocabularies are built from skeleton data alone, severing the link to the visual domain.
To address these challenges, we introduce Superman, a unified framework that bridges visual perception with temporal, skeleton-based motion generation.
Our solution is twofold. First, to overcome the modality disconnect, we propose a Vision-Guided Motion Tokenizer. Leveraging the natural geometric alignment between 3D skeletons and visual data, this module pioneers robust joint learning from both modalities, creating a unified, cross-modal motion vocabulary.
Second, grounded in this motion language, a single, unified MLLM architecture is trained to handle all tasks. This module flexibly processes diverse, temporal inputs, unifying 3D skeleton pose estimation from video (perception) with skeleton-based motion prediction and in-betweening (generation).
Extensive experiments on standard benchmarks, including Human3.6M, demonstrate that our unified method achieves state-of-the-art or competitive performance across all motion tasks.
This showcases a more efficient and scalable path for generative motion analysis using skeletons.
\vspace{-2em}
\end{abstract}

%% file: sec_cameraready/1_intro.tex
\section{Introduction}
\label{sec:intro}

\begin{table}[t]
    \setlength\tabcolsep{0.7mm}
    \centering
    \scriptsize
    \caption{Comparison with existing traditional and LLM/MLLM-based models for perceiving and/or generating human poses}
    \vspace{-0.8em}
    \begin{tabular}{l|c@{}c|c|c|c}
    \toprule
    & \multicolumn{2}{c|}{\textbf{Perception}} &\textbf{Generation} & \textbf{Tokenizer} & \textbf{Repr.}\\
    & vision & motion & motion & visual guidance \\    \midrule
    \multicolumn{6}{c}{\textit{Traditional Multi-Task Models}} \\   \midrule
    MotionBERT~\cite{zhu2023motionbert} & \textcolor{red}{$\times$} & \checkmark & \textcolor{red}{$\times$} & N/A & skeleton\\
    SiC~\cite{wang2024sic} & \textcolor{red}{$\times$} & \checkmark & \textcolor{red}{$\times$}  & N/A & skeleton\\
    HiC~\cite{liu2025human} & \textcolor{red}{$\times$} & \checkmark & \textcolor{red}{$\times$}  & N/A & skeleton\\ \midrule
    \multicolumn{6}{c}{\textit{LLM / MLLM-based Models}} \\   \midrule
    MotionLLM~\cite{chen2024motionllm}  & video & \checkmark & \textcolor{red}{$\times$}  & \textcolor{red}{$\times$} & SMPL\\
    MotionGPT~\cite{jiang2023motiongpt}  & \textcolor{red}{$\times$} & \checkmark & \checkmark & \textcolor{red}{$\times$} & SMPL\\
    MotionGPT3~\cite{zhu2025motiongpt3}  & \textcolor{red}{$\times$} & \checkmark & \checkmark & \textcolor{red}{$\times$} & SMPL\\
    LocLLM~\cite{wang2024locllm}& image & \textcolor{red}{$\times$} & \textcolor{red}{$\times$} & \textcolor{red}{$\times$}  & \textcolor{red}{$\times$} \\
    PoseLLaVA~\cite{feng2025posellava}& image & \textcolor{red}{$\times$} & \textcolor{red}{$\times$} & \textcolor{red}{$\times$}  & SMPL \\
    UniPose~\cite{li2025unipose}    & image & \textcolor{red}{$\times$} & \textcolor{red}{$\times$} & \textcolor{red}{$\times$}  & SMPL \\
    \midrule
    Superman & video/image & \checkmark & \checkmark & \checkmark & skeleton\\
    \bottomrule
    \end{tabular}
    \label{tab:summarize and compare different models}
    \par
    \begin{minipage}{\columnwidth}
    \textit{\footnotesize 
    ~``Motion'' refers to a temporal consecutive sequence of poses.
    }
    \end{minipage}
    \vspace{-2.5em}
\end{table}

Human motion analysis is a cornerstone of computer vision and robotics, with critical applications in human-computer interaction, augmented reality, and autonomous systems ~\cite{zhang2020distribution,andriluka2018posetrack,liu2018pem,zhang2024pose}. Several tasks such as 3D pose estimation \cite{zhang2020distribution,liu2018pem,zhang2024pose}, motion prediction~\cite{cui2021towards,wang2024dynamic,wang2024gcnext,fang2023explore}, and in-betweening~\cite{zhou2020generative,kaufmann2020convolutional,hernandez2019human,harvey2020robust} form the core of this field. 
Traditionally, these tasks have been tackled by highly specialized models, each optimized for a single objective. 
While this focused approach has yielded significant success, it has led to a fragmented and inflexible ecosystem.

The recent paradigm shift driven by Multi-modal Large Language Models (MLLMs) offers a compelling path toward unification. However, a fundamental disconnect persists, cleaving the field into two distinct camps. 
On one side, Multi-modal Large Language Models (MLLMs) like MotionLLM \cite{chen2024motionllm} and LLaVA-Pose \cite{zhang2025llava} excel at understanding motion. 
They can reason about poses from raw video and then generate textual descriptions, but lack the capacity to generate new, plausible human poses. 
On the other hand, specialized generative models like MotionGPT \cite{jiang2023motiongpt} excel at generating motion from text but struggle to process or perceive raw visual inputs. 
This ``read-only'' vs. ``write-only'' dichotomy forces researchers to use disparate architectures for tasks that are inherently interconnected. 
Moreover, constraint on input modality is also a limitation. 
Many generative methods operate only on abstract text or single images (\textit{e.g.,} ChatPose \cite{feng2024chatpose}, PoseLLaVA~\cite{feng2025posellava}, UniPose \cite{li2025unipose}), ignoring the rich temporal dynamics of video. 
Conversely, video-processing models such as SKI Models~\cite{sinha2025ski} are typically confined to understanding, not perception, i.e., generating textual results instead of human poses.
This fragmentation not only hinders research efficiency but also prevents models from leveraging the synergistic relationship between perception and generation.
This leaves a critical gap: the absence of a single, unified model that can fluidly process \textit{multi-modal inputs}, from \textit{full videos} for pose estimation to skeleton sequences for prediction, and generate coherent, structured 3D motion as output.

To unify perception with generation, we conceptualize motion as a universal language. 
Our core idea is to train a single MLLM to serve as a unified motion processor. 
The key to achieving this is a cross-modal motion vocabulary. \uline{Existing methods, however, build this vocabulary only from skeletal data, severing the link to the visual domain.}

\begin{figure*}[t]
    \centering
    \includegraphics[width=0.95\textwidth]{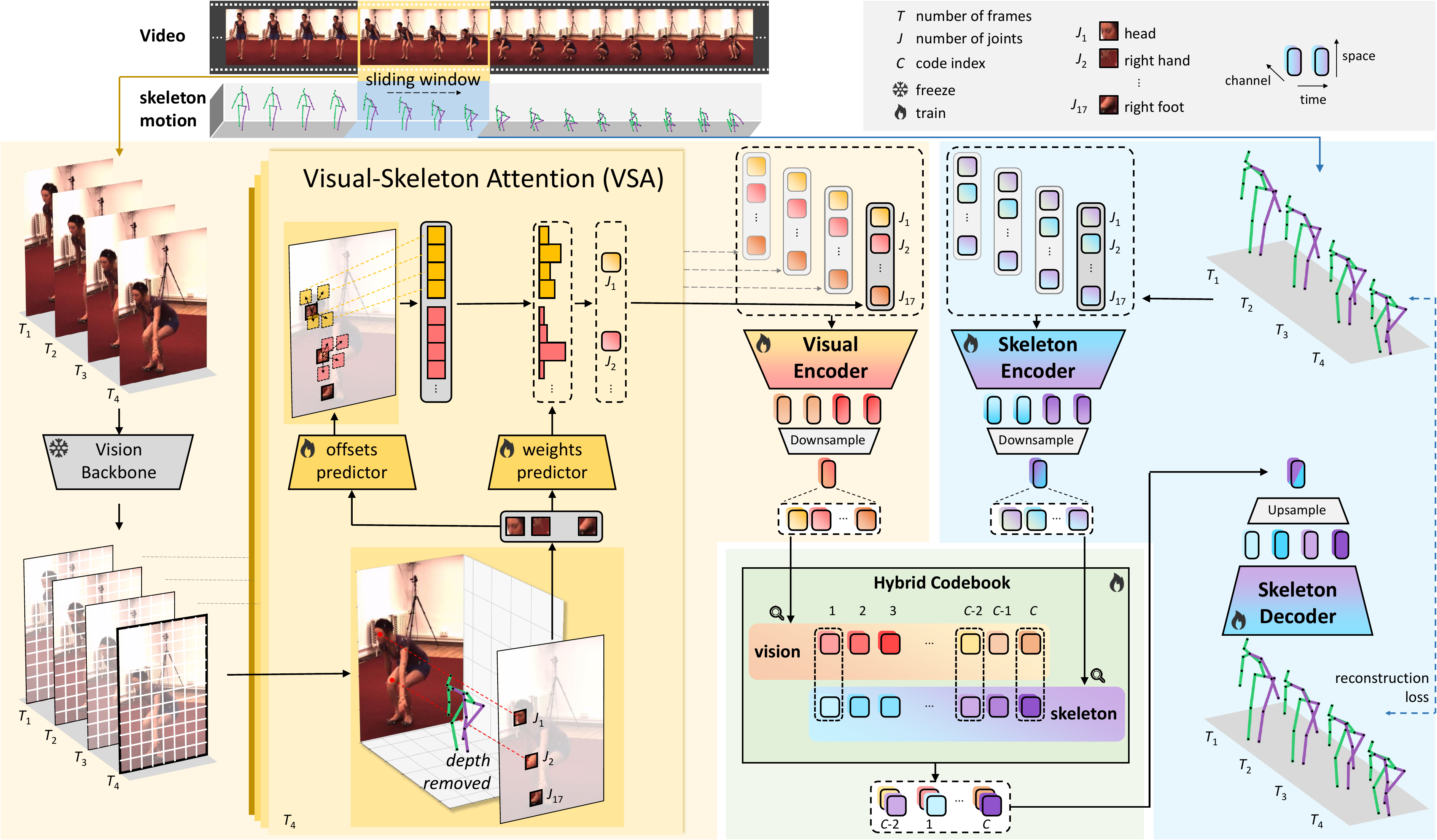}
    \vspace{-1em}
    \caption{
        \textbf{Architecture of our Vision-Guided Motion Tokenizer (VGMT).}
        VGMT creates discrete motion vocabulary by jointly fusing information from two modalities.
        A Skeleton Encoder ($E_s$) captures geometry while a Visual-Skeleton Attention (VSA) module and a subsequent Visual Encoder ($E_v$) ground the pose in visual features.
        The fused representation is quantized against a learnable hybrid codebook, and a decoder reconstructs the 3D poses.
    }
    \label{vqvae}
    \vspace{-1em}
\end{figure*}

To address this, we introduce \textbf{Superman}, a unified, multi-task, and multi-modal generative framework that reformulates these disparate challenges as a conditional sequence generation problem. Our approach has two key advantages: 1) it unifies perception and generation in a single model, promoting knowledge sharing between tasks; 2) it is flexible with multi-modal inputs, processing video, sequential skeletons, and text within the same architecture.
Our cornerstone is the Vision-Guided Motion Tokenizer, a VQ-VAE~\cite{van2017neural} architecture building a universal, discrete pose vocabulary. Unlike prior works limited to skeleton data, our tokenizer introduces a \textbf{hybrid codebook} for a richer, cross-modal representation. Each token comprises paired visual and geometric prototypes. This structure guides quantization using both visual appearance from frames and 3D skeleton geometry, creating a powerful representation that intrinsically links visual evidence to motion semantics.

Grounded in this motion language, our MLLM flexibly processes diverse inputs within a single architecture. 
It can: \textbf{(1) Estimate 3D Pose} by translating a video into a sequence of pose tokens. \textbf{(2) Predict Motion} by auto-regressively completing a pose sequence. \textbf{(3) Perform In-betweening} by generating intermediate tokens between keyframes. 
As shown in Fig.~\ref{fig:teaser}, our unified design is validated by extensive experiments on standard benchmarks including Human3.6M~\cite{ionescu2013h36m} and 3PDW~\cite{von2018_3dpw}.
Superman achieves state-of-the-art performances across all tasks, even when compared with traditional multi-task models specializing in perception tasks. For instance, for 3D pose estimation on Human3.6M~\cite{ionescu2013h36m}, Superman achieves an 11.97\% improvement compared with state-of-the-art multi-task perception method~\cite{liu2025human}, and a 10.91\% improvement compared with state-of-the-art LLM/MLLM-based method~\cite{feng2025posellava}.
Additionally, Superman delivers strong generalization performance. Trained exclusively on Human3.6M, Superman generalizes well to unseen data, namely, 3DPW~\cite{von2018_3dpw}, outperforming existing methods. 
Our contributions are threefold:

\begin{itemize}
    \item We propose a unified generative framework that leverages a single MLLM for multi-task, multi-modal human motion analysis, bridging the gap between motion perception and generation.
    \item We introduce a novel Vision-Guided Motion Tokenizer that creates a robust, cross-modal pose vocabulary. By designing a hybrid codebook where each token is a bimodal entity comprising paired visual and geometric prototypes, we ensure that the quantization process is simultaneously guided by both appearance features from video and the geometric structure of 3D skeletons.
    \item We demonstrate that our unified method achieves state-of-the-art across diverse tasks, validating the efficacy and scalability of our approach.
\end{itemize}

%% file: sec_cameraready/2_related_work.tex
\section{Related Work}
\label{sec:related_work}

\paragraph{Traditional Human Perception Models.}
Traditional models have evolved from classical methods to deep learning approaches using CNNs~\cite{toshev2014deeppose,pavllo20193d}, GCNs~\cite{martinez2017simple,wang2024gcnext,wang2024dynamic}, and Transformers~\cite{zheng2021poseformer}. However, traditional multi-task methods~\cite{liu2024point,fang2023explore,wang2024sic,liu2025human,zhu2023motionbert} often are limited in the scope of tasks, often focusing on only perception and not generation capabilities. In contrast, our work integrates multiple human perception and generation tasks into a unified generative framework based on MLLMs.
\vspace{-2mm}

\paragraph{Human Motion as a Language.}
A groundbreaking shift in motion analysis involves representing continuous human motion as a sequence of discrete tokens, a paradigm pioneered by early works in gesture synthesis~\cite{jiang2023motiongpt}. A common technique for this is using a Vector Quantized Variational Autoencoder (VQ-VAE) to learn a ``codebook'' of atomic poses~\cite{ding2025mtvcrafter}. As summarized in Tab.~\ref{tab:summarize and compare different models}, recent models have framed tasks like motion generation and prediction as language modeling problems~\cite{jiang2023motiongpt, zhang2024pose}. However, their scope is often limited to generating from non-visual modalities, failing to ground the motion language in raw visual input. Our work builds directly upon this paradigm but makes a crucial extension by integrating vision into the tokenization process, creating a visually grounded motion language.
\vspace{-4mm}

\paragraph{Multi-modal Large Language Models for Pose Perception and Generation.}
The remarkable success of Multi-modal Large Language Models (MLLMs), which integrate powerful vision encoders with LLMs for complex visual reasoning~\cite{liu2023visual,ye2023mplug,dai2023instructblip}, has inspired their application to human-centric tasks.
In the context of human motion, models such as MotionLLM~\cite{chen2024motionllm} and LLaVA-Pose~\cite{zhang2025llava} have demonstrated the ability to comprehend and answer questions about human actions and poses depicted in images and videos. 
They effectively connect the visual and semantic aspects of human motion. 
However, a significant limitation of these models is their primary focus on perception. They can interpret and describe poses, but they are not architecturally designed to generate new, structured motion sequences. 
Recent works, such as ChatPose~\cite{feng2024chatpose} and UniPose~\cite{li2025unipose}, have begun to bridge this gap by enabling pose generation conditioned on both visual and textual inputs. These models, however, typically operate on static, single-frame images. 
Our work distinguishes itself by being a fully versatile framework that not only perceives motion from videos but is also a powerful generative model capable of multiple synthesis tasks, thereby unifying the reading and writing of motion within a single, coherent architecture.

\begin{figure}[t]
    \centering
    \includegraphics[width=0.99\columnwidth]{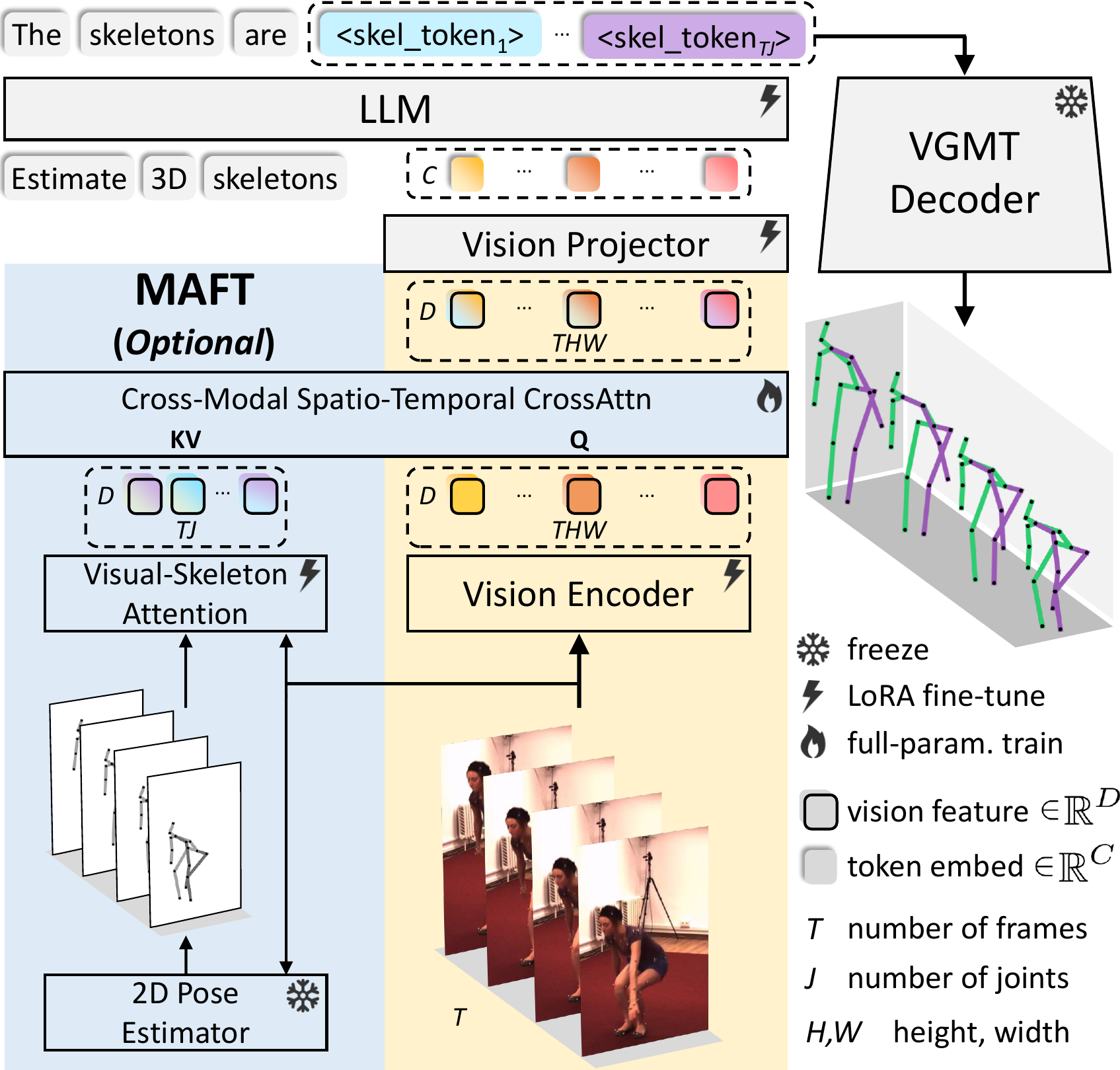}
    \vspace{-0.5em}
    \caption{
    \textbf{Network architecture and training paradigm.} Superman fine-tune a single LLM to integrate information from text, video, and 3D skeleton modalities. Optionally, a \textbf{Motion-Aware Fine-Tuning (MAFT)} module can be integrated. With $<$0.2\% extra parameters, MAFT enhances motion perception by enabling cross-video-motion fusion, leading to substantial improvement on tasks with visual input, as validated by the experiment results.
    }
    \label{network}
   \vspace{-2mm}
\end{figure}

%% file: sec_cameraready/3_method.tex
\section{Method}
\label{sec:methodology}

Our method unifies multi-task human motion analysis within a conditional sequence generation framework.
It involves two key stages: \textbf{(1) a novel Vision-Guided Motion Tokenizer (Section \ref{sec:tokenizer})} that converts continuous, high-dimensional motion into a discrete sequence of semantic tokens, and \textbf{(2) a Multi-modal Large Language Model (MLLM) (Section \ref{sec:llm_modeling})} that autoregressively models these token sequences to perform our diverse set of tasks. The overall network architecture is depicted in Fig.~\ref{network}.

\subsection{Vision-Guided Motion Tokenizer}
\label{sec:tokenizer}
The cornerstone of our approach is a discrete vocabulary for continuous human motion.
Unlike prior tokenizers that rely on skeletons alone~\cite{jiang2023motiongpt}, we design a Vision-Guided Motion Tokenizer, built upon a VQ-VAE, which is explicitly trained to fuse \textit{visual appearance} and \textit{3D skeletal geometry}. This ensures each token is grounded in both modalities. The architecture is illustrated in Fig.~\ref{vqvae}.

\begin{table*}[tp]
    \centering
    \renewcommand\arraystretch{1.1}
    \setlength\tabcolsep{1.3mm}
    \caption{
    \textbf{Comparison of our model with traditional multi-task models and LLM / MLLM-based models} on three human motion tasks: pose estimation (PE), motion prediction (MP), and motion in-betweening (MIB) on Human3.6M~\cite{ionescu2013h36m}.
    All tasks are evaluated using Mean Per Joint Position Error (MPJPE) in millimeter, averaged over all test data, where lower is better.
    ``$T$'' means how many frames the model inputs and outputs. (``$T$=1'' means the video has to be processed image-by-image.) ``N/A'' means the model is not applicable to the task. For example, single-frame-only models~\cite{wang2024locllm,li2025unipose,feng2025posellava} cannot do MP or MIB; LLM-based models~\cite{jiang2023motiongpt,zhu2025motiongpt3} cannot do PE.
    }
    \vspace{-0.5em}
    \begin{tabular}{l l c |ll| llll| lll}
        \toprule
        \multirow{2}{*}{Models}  & \multirow{2}{*}{Venue} & \multirow{2}{*}{$T$}   &\multicolumn{2}{c|}{\textbf{PE}$^\ddagger$} &\multicolumn{4}{c|}{\textbf{MP}$^\dagger$} &\multicolumn{3}{c}{\textbf{MIB}$^\dagger$} \\
            &   &  & N-MPJPE & MPJPE    & Avg & 80ms & 160ms &320ms      & Avg  & mid & last \\\midrule
        \multicolumn{12}{c}{\textit{Traditional Multi-Task Models}} \\\midrule
        MotionBERT~\cite{zhu2023motionbert} & ICCV'23 &16   & 47.07& 56.70 & 29.94  & 18.67        & 26.82 & 50.33 & 42.37  & 44.86        & 53.16 \\
        Skeleton-in-Context~\cite{wang2024sic} & CVPR'24 &16   & 45.26& 55.57 & 33.42  & 21.09        & 30.48 & 54.62 & 31.27  & 36.66        & 36.84 \\
        Human-in-Context~\cite{liu2025human} & Arxiv'25 &16   & 44.77& 53.86 & 26.66  & 14.36        & 23.58 & 47.34  & 31.13  & 37.01        & 35.49\\
        \textbf{Superman (w/ MAFT)} & Ours'26 &16 & \textbf{39.41}& \textbf{51.61}   & \textbf{26.13}$^\dagger$  & \textbf{13.70}        & \textbf{23.30} & \textbf{44.90} & \textbf{30.61}$^\dagger$  & \textbf{35.99}        & \textbf{35.13} \\\midrule
        \multicolumn{12}{c}{\textit{LLM / MLLM-based Models}} \\\midrule
        LocLLM~\cite{wang2024locllm} & CVPR'24 &1   & 49.32& 62.19 &\multicolumn{4}{c|}{\cellcolor{mygray}N/A} &\multicolumn{3}{c}{\cellcolor{mygray}N/A} \\
        UniPose~\cite{li2025unipose} & CVPR'25 &1  & 65.24& 106.96 &\multicolumn{4}{c|}{\cellcolor{mygray}N/A} &\multicolumn{3}{c}{\cellcolor{mygray}N/A} \\
        PoseLLaVA~\cite{feng2025posellava} & AAAI'25 &1   & 50.40& 62.43 &\multicolumn{4}{c|}{\cellcolor{mygray}N/A} &\multicolumn{3}{c}{\cellcolor{mygray}N/A} \\
        MotionGPT~\cite{jiang2023motiongpt} & NeurIPS'23 &16 & \multicolumn{2}{c|}{\cellcolor{mygray}N/A} & 48.81  & 42.99        & 49.27 & 55.31 & 52.96  & 56.07        & 66.45 \\
        MotionGPT3~\cite{zhu2025motiongpt3} & ICLR'26 &16 & \multicolumn{2}{c|}{\cellcolor{mygray}N/A} & 42.30  & 37.25        & 42.70 & 47.94 & 47.66  & 50.47        & 59.81 \\
        \textbf{Superman} & Ours'26 &16& \textbf{44.90} & \textbf{61.39}   & \textbf{26.13}$^\dagger$  & \textbf{13.70}        & \textbf{23.30} & \textbf{44.90} & \textbf{30.61}$^\dagger$  & \textbf{35.99}        & \textbf{35.13} \\
        \midrule
    \end{tabular}
    \label{tab:main results, all}
    \par
    \begin{minipage}{\textwidth}
    \textit{\footnotesize 
    $\ddagger$ Traditional models adopt ``RGB$\to$2D$\to$3D'' paradigm, leveraging a fixed 2D pose estimator; LLM/MLLM-based models adopt ``RGB$\to$3D'' paradigm.
    $\dagger$ For MP and MIB, all models take only 3D poses as input. Therefore, our MAFT module (which enhances vision) is deactivated, hence the identical results for Superman and Superman (w/ MAFT) on these two tasks.
    }
    \end{minipage}
    \vspace{-0.5em}
\end{table*}

\noindent
\textbf{Dual-Stream Encoding.}
Given a video clip $\mathcal{V} \in \mathbb{R}^{F \times H \times W \times 3}$ and its corresponding 3D pose sequence $\mathbf{X} \in \mathbb{R}^{F \times N \times 3}$, our tokenizer employs two parallel streams.

\noindent\textit{1) Visual Encoding Stream:} This stream extracts pose-centric visual features. We first use a vision backbone (e.g., HRNet~\cite{sun2019deep}) to obtain feature maps $\mathcal{F}_f$ for each frame. For each joint $j$ at frame $f$, its projected 2D location $\mathbf{p}_{j,f}$ serves as a reference point to sample an initial query feature $\mathbf{q}_{j,f}$ from $\mathcal{F}_f$. To enhance feature representation and robustness to occlusions, we introduce a Visual-Skeleton Attention (VSA) module. The VSA adaptively aggregates features by predicting sampling offsets and aggregation weights from the query. The visual feature for the joint is computed as:
\begin{equation}
    \mathbf{v}_{j,f} = \text{VSA}(\mathbf{q}_{j,f}, \mathcal{F}_f, \mathbf{p}_{j,f}),
\end{equation}
yielding a sequence of contextually-aware visual features $\mathbf{V} \in \mathbb{R}^{F \times N \times D}$.

\noindent\textit{2) Skeletal Encoding Stream:} This stream captures the motion's intrinsic geometry. The input pose sequence $\mathbf{X}$ is passed through a series of lightweight 2D convolutions operating on the joint-temporal grid. This effectively models spatio-temporal kinematics and maps the coordinates into a latent space, producing skeletal features $\mathbf{S} \in \mathbb{R}^{F \times N \times D}$.

\noindent
\textbf{Hybrid Codebook Quantization.}
To ensure each token represents a short \emph{motion pattern} rather than a static pose, we first temporally downsample the feature sequences $\mathbf{V}$ and $\mathbf{S}$, yielding window-level representations $\mathbf{z}^{v}_{w}$ and $\mathbf{z}^{s}_{w}$ for each time window $w$.
We then introduce a \emph{hybrid} codebook that pairs visual and geometric prototypes: $\mathcal{C} = \{(\mathbf{c}^{v}_k, \mathbf{c}^{s}_k)\}_{k=1}^{K}$, where $\mathbf{c}^{v}_k, \mathbf{c}^{s}_k \in \mathbb{R}^{D}$. For each window $w$, $k_w$ is the token index that minimizes the joint distance:
\begin{equation}
    k_w = \underset{k}{\arg\min} \left( \|\mathbf{z}^{v}_{w} - \mathbf{c}^{v}_k\|_2^2 + \|\mathbf{z}^{s}_{w} - \mathbf{c}^{s}_k\|_2^2 \right).
\end{equation}
This hybrid selection, based on minimizing the joint Euclidean distance, grounds the discretization process in both vision and geometry. The selected code vectors for the window are then $\hat{\mathbf{c}}^{v}_w=\mathbf{c}^{v}_{k_w}$ and $\hat{\mathbf{c}}^{s}_w=\mathbf{c}^{s}_{k_w}$.

\noindent
\textbf{Reconstruction and Training.} The skeletal code $\hat{\mathbf{c}}^{s}_w$ is upsampled and passed to a decoder to reconstruct the 3D poses $\hat{\mathbf{X}}_{w}$. The tokenizer is trained end-to-end with a VQ objective that combines reconstruction with modality-wise commitment losses:
\begin{footnotesize}
\begin{equation}
    \mathcal{L}_{\text{VQ}} = 
    \|\mathbf{X}_{w} - \hat{\mathbf{X}}_{w}\|_2^2
    + \beta_s \|\text{sg}[\mathbf{z}^{s}_{w}] - \hat{\mathbf{c}}^{s}_w\|_2^2
    + \beta_v \|\text{sg}[\mathbf{z}^{v}_{w}] - \hat{\mathbf{c}}^{v}_w\|_2^2,
\end{equation}
\end{footnotesize}
where $\text{sg}[\cdot]$ is the stop-gradient operator, and we set the commitment weights $\beta_s = 0.5$ and $\beta_v = 0.5$. This process effectively converts a continuous motion clip of $F$ frames into a discrete sequence of $T$ integer tokens $\mathbf{K}_{1:T}$. After training, the tokenizer provides a robust, visually-grounded motion vocabulary.

\subsection{Unified Multi-Task Modeling with LLM}
\label{sec:llm_modeling}

Our discrete motion tokenizer enables us to frame diverse motion tasks as a unified conditional sequence generation problem. 
We employ a decoder-only Multi-modal Large Language Model (MLLM), Qwen2.5-VL-7B~\cite{bai2025qwen2}, as our central sequence processor, which learns to predict motion tokens autoregressively.

\noindent\textbf{Connecting Tokenizer to Qwen-VL.}
Besides the standard tokenizer-MLLM integration paradigm~\cite{jiang2023motiongpt,li2025unipose}, we additionally design an optional light-weight module to inject skeletal geometry into the MLLM's visual stream, Motion-Aware Fine-tuning (MAFT). First, we enhance the ViT features using a Visual-Skeleton Attention (VSA) block. The ViT backbone extracts patch features $\mathbf{Z}_{\text{grid}}$. Concurrently, 2D joint projections serve as reference points for a multi-scale deformable sampler~\cite{zhu2020deformable} to aggregate pose-centric features $\mathbf{Z}_{\text{pose}}$. We then fuse this information using VSA, implemented as a cross-attention layer followed by an FFN, where grid tokens are queries and pose tokens are keys/values. The visual tokens $\hat{\mathbf{Z}}_{\text{grid}}$ are computed as:
\begin{equation}
    \hat{\mathbf{Z}}_{\text{grid}} = \text{VSA}(\mathbf{Z}_{\text{grid}}, \mathbf{Z}_{\text{pose}}).
\end{equation}
These enhanced visual tokens, along with any textual prompts, are fed into the LLM decoder, which generates the final sequence of motion tokens.

\begin{table*}[tp]
    \centering
    \renewcommand\arraystretch{1}
    \setlength\tabcolsep{1.5mm}
    \caption{
    \textbf{Action-specific N-MPJPEs} for pose estimation on Human3.6M~\cite{ionescu2013h36m}.
    } 
    \vspace{-1em}
    \footnotesize
    \begin{tabularx}{0.99\textwidth}{l l c *{9}{>{\centering\arraybackslash}X}}
        \midrule
        \multirow{2}{*}{Models}  & \multirow{2}{*}{Venue} & \multirow{2}{*}{$T$}  &\multicolumn{8}{c}{\textbf{Actions}} \\\cmidrule{4-11}
            &   &   & Sit   & SitDown       & Smoke & Photo & Wait  & Walk  & WalkDog       & WalkTwo \\\midrule
        \multicolumn{11}{c}{\textit{Traditional Multi-Task Models}} \\\midrule
        MotionBERT~\cite{zhu2023motionbert} & ICCV'23 &16 & 62.22       & 69.81 & 49.21 & 49.66 & 39.49 & 40.80 & 44.94 & 40.95 \\
        Skeleton-in-Context~\cite{wang2024sic} & CVPR'24 &16 & 59.34    & 65.89 & 48.70 & 48.11 & 37.28 & 38.52 & 42.12 & 38.97 \\
        Human-in-Context~\cite{liu2025human} & Arxiv'25 &16 & 59.95     & 65.21 & 47.81 & 47.02 & 36.37 & 38.17 & 41.76 & 37.31 \\
        \textbf{Superman (w/ MAFT)} & Ours'26 &16 & \textbf{47.12}     & \textbf{48.97} & \textbf{41.31} & \textbf{34.37} & 41.21 & \textbf{36.12} & \textbf{39.23} & 43.23 \\\hline
        \multicolumn{11}{c}{\textit{LLM / MLLM-based Models}} \\\midrule
        LocLLM~\cite{wang2024locllm} & CVPR'24 &1 & 64.48       & 68.76 & 50.96 & 55.16 & 42.55 & 39.11 & 49.96 & 43.10 \\
        UniPose~\cite{li2025unipose} & CVPR'25 &1 & 72.30       & 84.24 & 69.20 & 62.65 & 60.02 & 66.97 & 65.71 & 70.69 \\
        PoseLLaVA~\cite{feng2025posellava} & AAAI'25 &1 & 64.64 & 71.36 & 52.67 & 54.21 & 44.67 & 46.90 & 48.90 & 44.90 \\
        \textbf{Superman} & Ours'26 &16 & \textbf{62.68}  & 71.93 & \textbf{49.09} & \textbf{43.54} & \textbf{35.21} & \textbf{30.89} &\textbf{ 45.91} & \textbf{39.60} \\
        \bottomrule
    \end{tabularx}
    \vspace{-1em}
\label{tab:main results, pe, h36m}
\end{table*}

Our framework handles multiple tasks by varying the conditioning inputs:
\vspace{-4mm}
\paragraph{3D Pose Estimation from Video:}
Given a video clip $\mathbf{I}_{1:F}$, the MLLM is conditioned on its enhanced visual features $\hat{\mathbf{Z}}_{\text{grid}}$ to generate the corresponding motion token sequence $\mathbf{K}_{1:T}$. The objective is to maximize the log-likelihood of the ground-truth sequence:
    \begin{equation}
        \mathcal{L}_{\text{est}} = \sum_{t=1}^{T} \log P(k_t | \mathbf{K}_{<t}, \hat{\mathbf{Z}}_{\text{grid}}).
    \end{equation}
The tokens are then decoded to obtain the final 3D poses.
\vspace{-4mm}
\paragraph{Motion Prediction:}
Given a historical pose sequence, tokenized as $\mathbf{K}_{1:T'}$, the task is to predict the future. The LLM autoregressively generates subsequent tokens $\mathbf{K}_{T'+1:T}$ conditioned on the past:
    \begin{equation}
        \mathcal{L}_{\text{pred}} = \sum_{t=T'+1}^{T} \log P(k_t | \mathbf{K}_{<t}).
    \end{equation}
\vspace{-4mm}
\paragraph{Motion In-betweening:}
Given start and end tokens ($k_1$, $k_T$), we use a prompt like ``[START] $k_1$ [MIDDLE] $k_T$ [END]'', training the LLM to fill in the missing sequence corresponding to the [MIDDLE] segment.

Our framework's strength is a single LLM trained jointly on a mixture of task-specific formats. This multi-task strategy fosters a rich, unified representation, allowing it to act as a versatile motion processor.

\noindent\textbf{Training.}
Our framework is trained in two stages. First, the Vision-Guided Motion Tokenizer is independently trained, and its weights are frozen. Subsequently, the MLLM is trained on a mixture of all tasks using a single, unified objective: the standard autoregressive cross-entropy loss, summed across all task-specific samples in a batch. This joint strategy facilitates knowledge transfer between perception and generation.

\begin{table}[tp]
    \centering
    \renewcommand\arraystretch{1.1}
    \caption{
    \textbf{Comparison of results on generalization to unseen data.} Motion prediction (MP) and motion in-betweening (MIB) on 3DPW~\cite{von2018_3dpw} are reported. All models are trained on Human3.6M only, with 3DPW completely excluded from training.
    }
    \vspace{-1em}
    \scalebox{0.75}{
    \begin{tabular}{l  |lll| ll}
        \toprule
        \multirow{2}{*}{Models}    &\multicolumn{3}{c|}{\textbf{MP}} &\multicolumn{2}{c}{\textbf{MIB}} \\
              & Avg &80ms &320ms      & Avg  & mid \\\midrule
        \multicolumn{6}{c}{\textit{Traditional Multi-Task Models}} \\\midrule
        MotionBERT~\cite{zhu2023motionbert}   & 164.96    &137.89     & 200.14 & 123.05         & 148.51\\
        Skeleton-in-Context~\cite{wang2024sic}  & 140.71    &110.77     & 183.74& 103.97         & 127.68\\
        Human-in-Context~\cite{liu2025human}& 141.90     &112.53    & 183.62& 108.54         & 131.99\\\midrule
        \multicolumn{6}{c}{\textit{LLM / MLLM-based Models}} \\\midrule
        MotionGPT~\cite{jiang2023motiongpt} & 292.87    &257.92     & 331.88 &234.39 &257.55 \\
        MotionGPT3~\cite{zhu2025motiongpt3} & 228.17    &190.83     & 259.49 &180.10 &204.71 \\
        \textbf{Superman (Ours)}  & \textbf{62.05} &\textbf{34.75} & \textbf{97.37} & \textbf{60.68}  & \textbf{63.35}\\
        \midrule
    \end{tabular}
    }
    \label{tab:main results, 3dpw}
    \vspace{-2mm}
\end{table}

%% file: sec_cameraready/4_exp.tex
\section{Experiment}
\label{sec:experiments}

\noindent
\textbf{Implementation Details.}
Our framework is built upon Qwen2.5-VL-7B~\cite{bai2025qwen2} as the base MLLM.
We employ LoRA and, optionally, MAFT. The trainable parameters accounts for 12.67\% (w/o MAFT) and 12.87\% (w/ MAFT) of vanilla Qwen2.5-VL-7B.
Our Vision-Guided Motion Tokenizer has a vocabulary size of $K=8192$.
The model is trained on 2 NVIDIA H20 GPUs.

\noindent
\textbf{Datasets and Metrics.}
For evaluation, we utilize two large-scale 3D human motion datasets: Human3.6M~\cite{ionescu2013h36m} and 3DPW~\cite{von2018_3dpw}.
The models are trained on Human3.6M only, and tested on both datasets. The training set and the test set are in line with their respective splitting standards.
Tasks are evaluated using Mean Per Joint Position Error (MPJPE) as the main metric, in line with standard protocols.

\noindent
\textbf{Data Preprocessing.}
To unify formats, we first convert 3DPW's SMPL vertices to the Human3.6M skeleton format using a pre-trained matrix~\cite{bogo2016keep}. Then, both datasets follow the same processing pipeline. Specifically, for the X and Y spatial dimensions, we transform the 3D camera coordinates (in millimeters) into 2D image coordinates (in pixels), following~\cite{zhu2023motionbert}. For the depth (Z) dimension, we normalize the skeleton by setting the depth of the root joint (pelvis) to 0. The depths of all other joints are then converted to be relative to this root. Crucially, we do not translate the root joint to the coordinate origin (0, 0, 0) during either the training or testing phases. This approach retains the global positional information in the image coordinate space.

\begin{figure}[t]
    \centering
    \includegraphics[width=0.99\columnwidth]{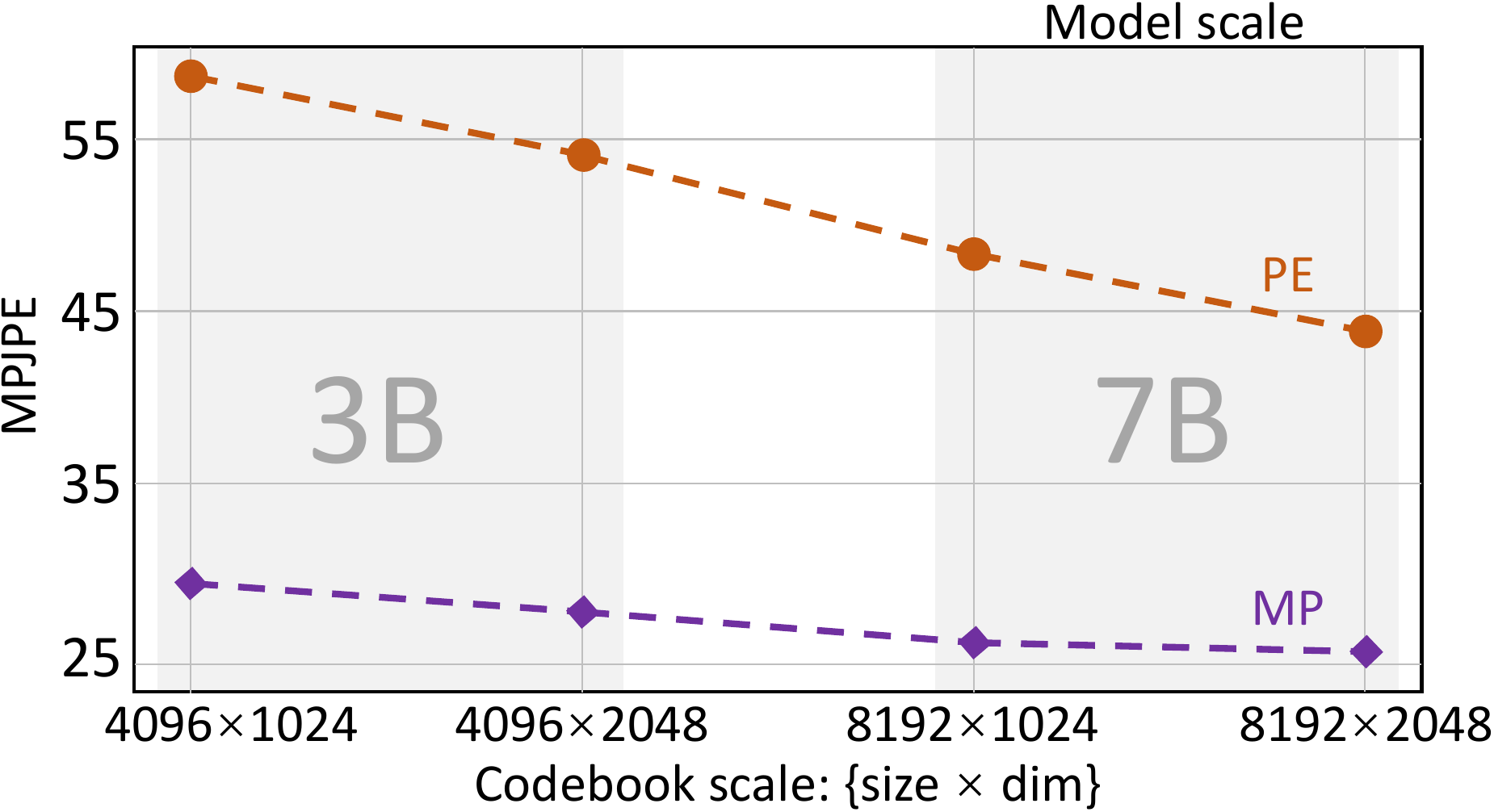} 
    \vspace{-1em}
    \caption{
        Scaling analysis of model and codebook parameters. Scaling up the model (3B$\to$7B) and the codebook consistently reduces (N-)MPJPE for both pose estimation (PE) and motion prediction (MP) tasks, demonstrating the method's ability to leverage larger capacity for improved accuracy.
    }
    \label{fig:exp abl scale}
    \vspace{-2mm}
\end{figure}

\begin{figure}[t]
    \centering
    \includegraphics[width=0.99\columnwidth]{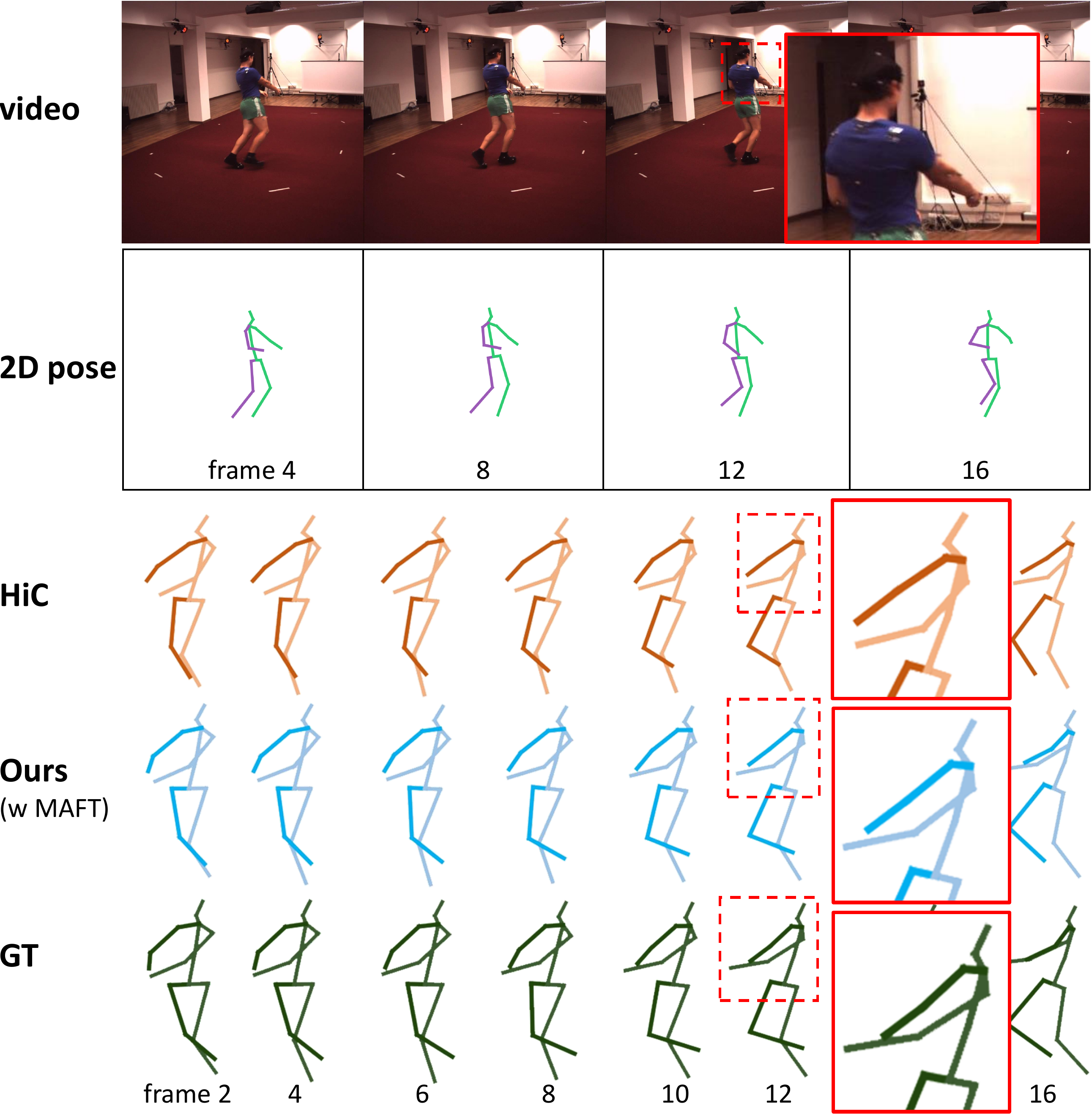} 
    \vspace{-0.5em}
    \caption{
        Qualitative results for pose estimation on Human3.6M. Our method with MAFT is compared with HiC~\cite{liu2025human}, the current SoTA. Both methods use video and 2D poses as inputs and leverages a fixed 2D pose estimator.
    }
    \label{fig:viz pe}
\end{figure}

\subsection{Main Results}
\label{sec:mian_results}

Our unified model, Superman, was evaluated on three human motion tasks: 3D Pose Estimation (PE), Motion Prediction (MP), and Motion In-Betweening (MIB). As shown in Tabs.~\ref{tab:main results, all},~\ref{tab:main results, pe, h36m}, and~\ref{tab:main results, 3dpw}, our model demonstrates state-of-the-art (SOTA) performance on both the standard benchmark (Human3.6M~\cite{ionescu2013h36m}) and generalization tests (3DPW~\cite{von2018_3dpw}).

\noindent
\textbf{Performance on Human3.6M.}
In the comprehensive evaluation (Tab.~\ref{tab:main results, all}), our model surpassed all baselines under both paradigms. Compared to traditional multi-task models (which use a ``RGB$\to$2D$\to$3D'' paradigm), Superman achieved the lowest errors of all models.
Compared to LLM/MLLM-based models (which use a ``RGB$\to$3D'' paradigm), Superman also secured the best performance with the lowest MPJPE/N-MPJPE.
In the action-specific N-MPJPE breakdown (Tab.~\ref{tab:main results, pe, h36m}), Superman achieved the lowest error in most of the action categories.

\begin{figure}[t]
    \centering
    \includegraphics[width=0.99\columnwidth]{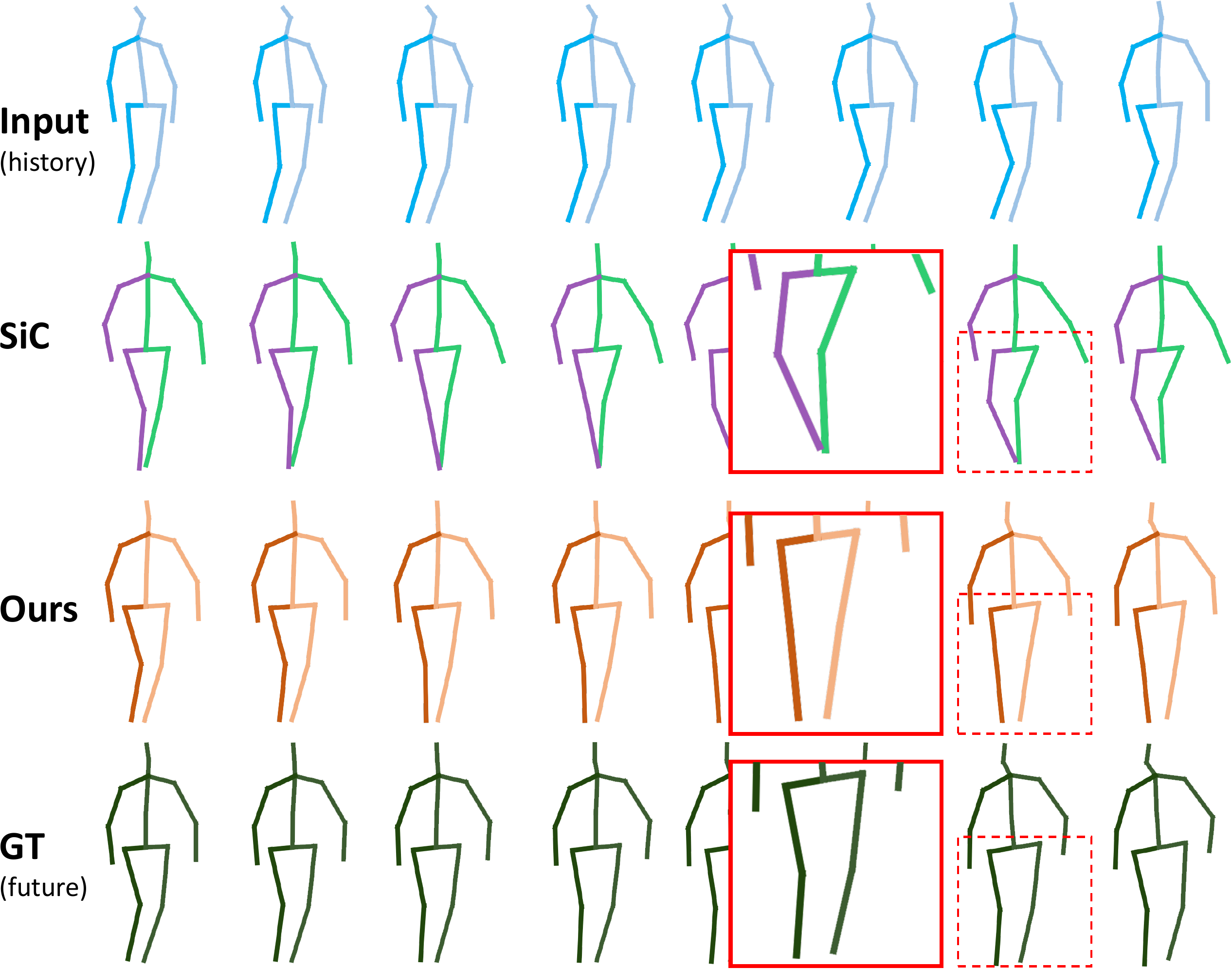} 
    \vspace{-0.5em}
    \caption{
        Qualitative results for generalizing to motion prediction on 3DPW (unseen dataset). Our method is compared with SiC~\cite{liu2025human}, the current SoTA on the task.
    }
    \label{fig:viz mp 3dpw}
\end{figure}

\noindent
\textbf{Generalization on 3DPW.}
To evaluate the model's ability to generalize to unseen data, we conduct a zero-shot test on the 3DPW dataset; all models are trained only on Human3.6M and tested on 3DPW.
As shown in Tab.~\ref{tab:main results, 3dpw}, Superman demonstrates exceptional generalization on both motion tasks, outperforming all baseline models by a significant margin. 
Fig.~\ref{fig:viz mp 3dpw} provides qualitative validation for this, showing our model generating more accurate future poses than the current SoTA baseline~\cite{wang2024sic}.
On motion in-betweening (MIB), our model achieved an average error of 60.68, again far surpassing the best traditional model and other LLM/MLLM models.
These results confirm that our model not only excels on the benchmark dataset but also possesses superior robustness and generalization capabilities when handling unseen ``in-the-wild'' data.

\noindent
\textbf{Computational Efficiency.}
Tab.~\ref{tab:efficiency_analysis} breaks down the efficiency metrics for both VGMT and MLLM components. Specifically, it details GFLOPs, parameter counts, and latency for the vision backbone, vision encoder, and VSA. 
While the large-scale Qwen2.5-VL-7B dominates overall resource consumption, accounting for 99.7\% of the total GFLOPs and 98.6\% of the total latency, our newly introduced components, such as MAFT and VSA, exhibit remarkably low overhead. Collectively, these proposed modules account for less than 0.03\% of the total computation cost, demonstrating the exceptional efficiency and practical scalability of our architecture.

\subsection{Visualization}

\begin{figure}[t]
    \centering
    \includegraphics[width=0.99\columnwidth]{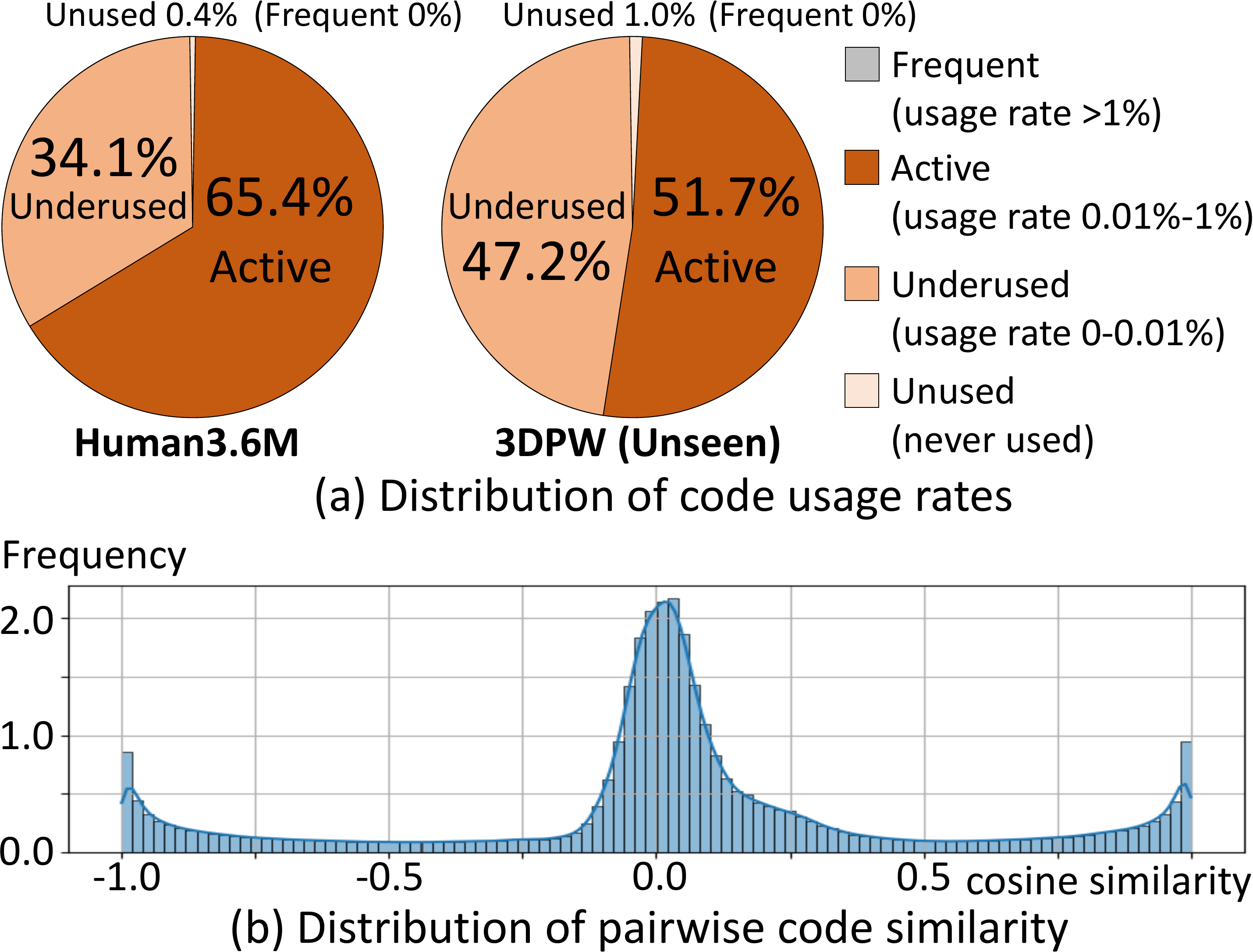}
    \vspace{-0.5em}
    \caption{
        Quantitative analysis of the VQ-VAE codebook. (a) Up to 65.4\% of the codes remain active during inference, indicating efficient usage of the encoding space. (b) The cosine similarity of most code pairs is close to 0, confirming the ability to learn a discrete latent space with highly decorrelated representations.
    }
    \label{fig:exp abl vqvae pie}
    \vspace{-0.5em}
\end{figure}

\paragraph{Analysis of the VQ-VAE Codebook.} To evaluate the efficiency of our codebook utilization, we conducted a quantitative analysis. As shown in Fig.~\ref{fig:exp abl vqvae pie}~(a), the codes are categorized into four usage frequency levels: Frequent (usage rate $>$1\%), Active (0.01\%-1\%), Underused (0-0.01\%), and Unused (never used). On the Human3.6M dataset, up to 65.4\% of the codes remain Active, indicating efficient utilization of the encoding space. This high efficiency is maintained even on the unseen 3DPW dataset, which shows 51.7\% Active codes and only 1.0\% Unused codes. Furthermore, we analyzed the codebook’s latent space by computing pairwise cosine similarities. The distribution in Fig.~\ref{fig:exp abl vqvae pie}~(b) shows that the similarity of most code pairs is close to 0, confirming the model's ability to learn a discrete latent space with highly decorrelated representations.
\vspace{-4mm}
\paragraph{Qualitative Results for 3D Pose Estimation.} Figure~\ref{fig:viz pe} presents qualitative visualizations for 3D pose estimation on Human3.6M. The figure provides an intuitive comparison of our model (``Ours'' shown in blue) against the baseline method HiC~\cite{liu2025human} (``HiC'' shown in orange) and the Ground Truth (``GT'' shown in dark green). Visually, the 3D skeleton sequence generated by our method shows a high degree of fidelity to the Ground Truth in both shape and motion, while the baseline method exhibits noticeable visual discrepancies.

\begin{table}[t]
    \vspace{-1em}
    \centering
    \setlength\tabcolsep{1.2mm}
    \captionof{table}{Computation efficiency.}
    \vspace{-1em}
    \begin{tabular}{lrrr}
    \toprule
    Module & GFLOPs & params & latency  \\
      &   &  (M) & (ms)  \\
    \midrule
    \multicolumn{4}{c}{\textit{Vision-Guided Motion Tokenizer (VGMT)}} \\\midrule
    vision backbone \textcolor{gray}{$^{\text{\scriptsize\faLock}}$} & 31.375 & 28.53 & 43.7  \\
    vision encoder  & 2.468  & 17.12 & 2.7   \\
    VSA             & 0.001  & 0.11  & 0.1   \\
    \midrule
    \multicolumn{4}{c}{\textit{MLLM}} \\\midrule
    MAFT & 5.01 & 25.45 & 0.3 \\
    Qwen2.5-VL-7B   &     $\sim$129920.00   & $\sim$9605.00 & $\sim$3300.0   \\
    
    \bottomrule
    \end{tabular}
    \label{tab:efficiency_analysis}
    \par
    \begin{minipage}{\columnwidth}
    \textit{\footnotesize 
    ~~Vision backbone remains frozen during training.
    }
    \end{minipage}
\end{table}

\subsection{Ablation Study and Analysis}
We conduct comprehensive ablation studies to validate the effectiveness of our framework's key design choices.

\vspace{-4mm}
\paragraph{Analysis of Model and Codebook Scaling.}
To validate the scalability of our framework, we investigate the impact of both the MLLM's parameter size and the codebook's capacity. As illustrated in Fig.~\ref{fig:exp abl scale}, we evaluate performance (MPJPE) by scaling the model from 3B to 7B parameters, while simultaneously increasing the codebook scale (in terms of both size and dimension).
The results show a clear and consistent trend: (1) Increasing the model size from 3B to 7B significantly reduces the error for both Pose Estimation (PE) and Motion Prediction (MP). (2) Within a given model size, increasing the codebook capacity (e.g., from $4096 \times 1024$ to $4096 \times 2048$ for the 3B model) also consistently lowers the MPJPE. This analysis confirms that our unified framework benefits from larger model and codebook parameters, demonstrating strong scalability and potential for future improvements.

\vspace{-4mm}
\paragraph{Effectiveness of the Vision-Guided Tokenizer.}
We validate our tokenizer design in Table~\ref{tab:ablation_tokenizer}. Results show that jointly learning from vision and geometry is essential. The Visual-Only model fails (22.5 mm error), and while the Skeleton-Only baseline~\cite{jiang2023motiongpt} is reasonable (7.7 mm), it is significantly outperformed by all fused configurations. Among these, our chosen balanced weighting ($\beta_s=0.5, \beta_v=0.5$) achieves the lowest reconstruction error (4.7 mm), proving superior to other weightings (5.3-6.8 mm) and validating our approach.

\begin{table}[t]
    \centering
    \caption{Ablation on tokenizer components and fusion weights.}
    \vspace{-1em}
    \begin{tabular}{lcc}
    \toprule
    Tokenizer Config. & Recon. Err. & PE (N-MPJPE)\\
    \midrule
    Visual-Only & 22.5 & 51.3 \\
    Skeleton-Only & 7.7 & 47.8 \\
    \midrule
    Fused: $\beta_s, \beta_v=0.3,0.7$ & 6.8 &45.8\\
    Fused: $\beta_s, \beta_v=0.7,0.3$ & 5.3 &45.1\\
    \rowcolor{mygray}Fused: $\beta_s, \beta_v=0.5,0.5$ & \textbf{4.7} & \textbf{44.9}\\
    \bottomrule
    \end{tabular}
    \label{tab:ablation_tokenizer}
    \vspace{-1em}
\end{table}

\vspace{-4mm}
\paragraph{Effectiveness of Unified Multi-Task Training.}
Effectiveness of Unified Multi-Task Training. Our second key design is the use of a single MLLM for joint multi-task training (as described in Sec.~\ref{sec:llm_modeling}). To validate this, we compare our Unified model against Specialized models, which use the same architecture but are trained exclusively on a single task. As shown in Tab.~\ref{tab:ablation_multitask}, our Unified model consistently outperforms its Specialized counterparts across all three tasks. The unified model achieves a PE of 44.9 (vs. 46.5), an MP of 26.1 (vs. 27.3), and an MiB of 30.6 (vs. 33.1). This strongly indicates positive knowledge transfer between tasks, as learning them jointly improves performance in all areas. This confirms our unified framework is not only efficient but also more effective by learning a richer, joint representation.

\begin{table}[t]
    \centering
    \caption{Ablation on the unified multi-task training strategy. We compare our single unified model against specialized models.}
    \scalebox{0.9}{
    \begin{tabular}{l|ccc}
    \toprule
    \textbf{Training Strategy} & \makecell{PE} & \makecell{MP} & \makecell{MIB} \\
    \midrule
    Specialized (Est. Only) & 46.5 & N/A & N/A \\
    Specialized (Pred. Only) & N/A & 27.3 & N/A \\
    Specialized (MiB. Only) & N/A & N/A & 33.1 \\
    \rowcolor{mygray}
    Unified & \textbf{44.9} & \textbf{26.1} & \textbf{30.6} \\
    \bottomrule
    \end{tabular}
    }
    \label{tab:ablation_multitask}
\end{table}

%% file: sec_cameraready/5_conclusion.tex
\section{Conclusion}
\label{sec:conclusion}

We propose a unified generative framework that bridges the gap between motion perception and generation. 
The core of our approach is a novel Vision-Guided Motion Tokenizer, which, unlike prior work, learns a robust cross-modal vocabulary by jointly fusing visual features and geometric data. 
Grounded in this motion language, our single MLLM architecture achieves state-of-the-art or competitive performance across diverse tasks, including 3D pose estimation, motion prediction, and motion in-betweening.
This work demonstrates that a unified, visually grounded generative model is an efficient, scalable paradigm for holistic human motion analysis.

\section{Acknowledgement}
This work was supported by National Natural Science Foundation of China (No. 62473007), Guangdong Outstanding Youth Fund (No. 2026B1515020015), Shenzhen Innovation in Science and Technology Foundation for The Excellent Youth Scholars (No. RCYX20231211090248064).

%% file: sec_cameraready/X_suppl.tex
\clearpage
\setcounter{page}{1}

\setcounter{section}{0}
\setcounter{table}{0}
\setcounter{figure}{0}
\setcounter{equation}{0}

\renewcommand{\thesection}{\Alph{section}}
\renewcommand{\thetable}{S\arabic{table}}
\renewcommand{\thefigure}{S\arabic{figure}}
\renewcommand{\theequation}{S\arabic{equation}}

\twocolumn[{%
    \renewcommand\twocolumn[1][]{#1}%
    
    \maketitlesupplementary
    \centering
    \captionsetup{type=figure} 

    \includegraphics[width=0.99\textwidth]{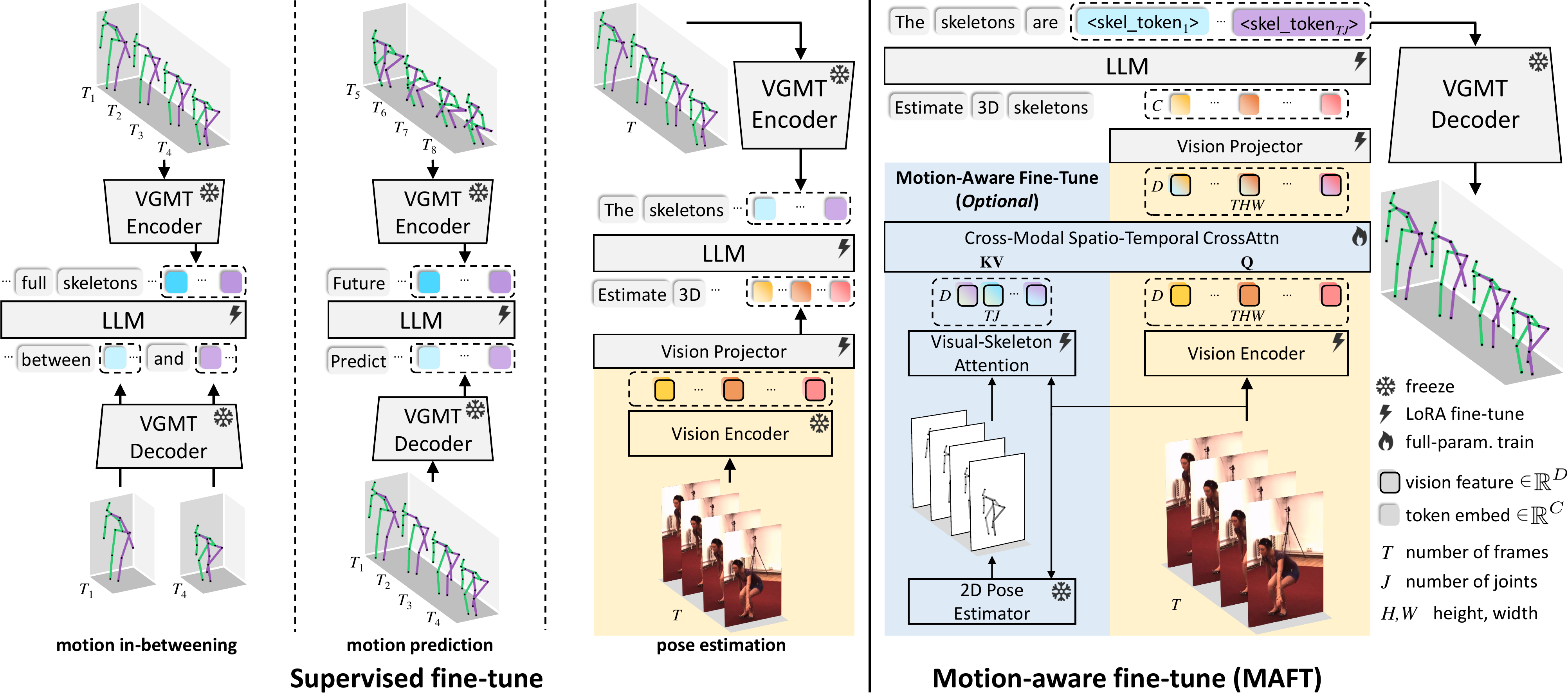}
    \caption{\textbf{Detailed Data Flow and Training Pipelines for Different Tasks.} 
    \textbf{Left:} For generation tasks like \textit{Motion In-betweening} and \textit{Motion Prediction}, the model operates in a "Skeleton-to-Skeleton" mode. Historical or keyframe skeletons are tokenized by the frozen VGMT Encoder, and the LLM autoregressively predicts the missing/future tokens.
    \textbf{Right:} For \textit{Pose Estimation}, the model operates in a "Video-to-Skeleton" mode. We compare the standard Supervised Fine-Tuning (SFT) pipeline against our proposed \textbf{Motion-Aware Fine-Tuning (MAFT)}. The MAFT pipeline explicitly injects skeletal geometry into the visual stream via the VSA module, enhancing feature alignment before LLM processing.}
    \label{fig:supp:viz fig_ft_stage}
    \vspace{1em}
}]

\section{Implementation Details}
    \label{sec:supp_chap4_impl}

\begin{table*}[h]
    \centering
    \caption{\textbf{Instruction Templates for Multi-Task Training.} We utilize specific text prompts to guide the MLLM for different tasks. The bottom section illustrates a concrete example of the actual input prompt (with special vision tokens) and the corresponding output response.}
    \label{tab:supp_prompts}
    \renewcommand{\arraystretch}{1.3} 
    \resizebox{\textwidth}{!}{
    \begin{tabular}{c|c|p{11cm}|c}
        \toprule
        \textbf{Task} & \textbf{Input Modality} & \multicolumn{1}{c|}{\textbf{Instruction Template (User Prompt)}} & \textbf{Output} \\
        \hline
        
        \multirow{3}{*}{\shortstack{3D Pose\\Estimation\\(Vid2Skel)}} & \multirow{3}{*}{Video} & 
        Please describe the motion of the person in the video \texttt{<video>} using skeleton tokens. Your response should be formatted as: & \multirow{3}{*}{\shortstack{Skeleton\\Tokens\\($K_{1:T}$)}} \\
        & & "Frame 1: torso: ... left\_arm: ... right\_arm: ... left\_leg: ... right\_leg: ... & \\
        & & Frame 2: ... " & \\
        \hline
        
        \multirow{3}{*}{\shortstack{Motion\\Prediction\\(SkelPred)}} & \multirow{3}{*}{\shortstack{History\\Skeleton}} & 
        Here's a motion sequence represented using skeleton tokens: \newline \texttt{<skeleton>} \newline Predict the future motion using skeleton tokens that have the same number of frames as the history motion. Your response should be formatted as: & \multirow{3}{*}{\shortstack{Future\\Skeleton\\Tokens}} \\
        & & "Future Frame 1: torso: ... left\_arm: ... right\_arm: ... left\_leg: ... right\_leg: ... & \\
        & & Future Frame 2: ... " & \\
        \hline
        
        \multirow{4}{*}{\shortstack{Motion\\In-betweening\\(MIB)}} & \multirow{4}{*}{\shortstack{Start/End\\Keyframes}} & 
        Here's a motion sequence with missing in-between frames. It contains only the start and end keyframes, represented using skeleton tokens: \newline \texttt{<skeleton>} \newline Generate the in-between frames to create a smooth transition between the provided keyframes. Your response should be formatted as a complete sequence: & \multirow{4}{*}{\shortstack{Intermediate\\Skeleton\\Tokens}} \\
        & & "Frame 1: ... left\_arm: ... right\_arm: ... left\_leg: ... right\_leg: ... & \\
        & & Frame 2: ... " & \\
        \hline
        
        \multicolumn{4}{l}{\cellcolor{gray!10}\textbf{Real Input Prompt Example (with Special Tokens):}} \\
        \multicolumn{4}{p{16cm}}{
        \small
        \textit{User:} Please describe the motion of the person in the video \texttt{<|vision\_start|><|video\_pad|>...<|video\_pad|><|vision\_end|>} using skeleton tokens. Your response should be formatted as: "Frame 1: torso: ... left\_arm: ... right\_arm: ... left\_leg: ... right\_leg: ... Frame 2: ... "
        } \\
        \hline
        
        \multicolumn{4}{l}{\cellcolor{gray!10}\textbf{Real Output Example (Body-Part Level Grouping):}} \\
        \multicolumn{4}{p{16cm}}{
        \small
        \textit{Assistant:} There are 4 frames in total. Here are the skeleton tokens for 5 body parts in each frame: \newline
        \texttt{Frame 1: torso: <skel\_1848><skel\_7294><skel\_188><skel\_5434><skel\_975>. left\_arm: <skel\_5563><skel\_6334><skel\_7382>. right\_arm: <skel\_1656><skel\_7242><skel\_7368>. left\_leg: <skel\_4964><skel\_445><skel\_426>. right\_leg: <skel\_2974><skel\_3963><skel\_1134>.} \newline
        \texttt{Frame 2: torso: <skel\_7234><skel\_7612><skel\_7832><skel\_2617><skel\_3466>. left\_arm: <skel\_2904><skel\_165><skel\_5270>. ...}
        } \\
        \bottomrule
    \end{tabular}
    }
\end{table*}

\begin{figure*}[t]
    \centering
    \includegraphics[width=0.9\textwidth]{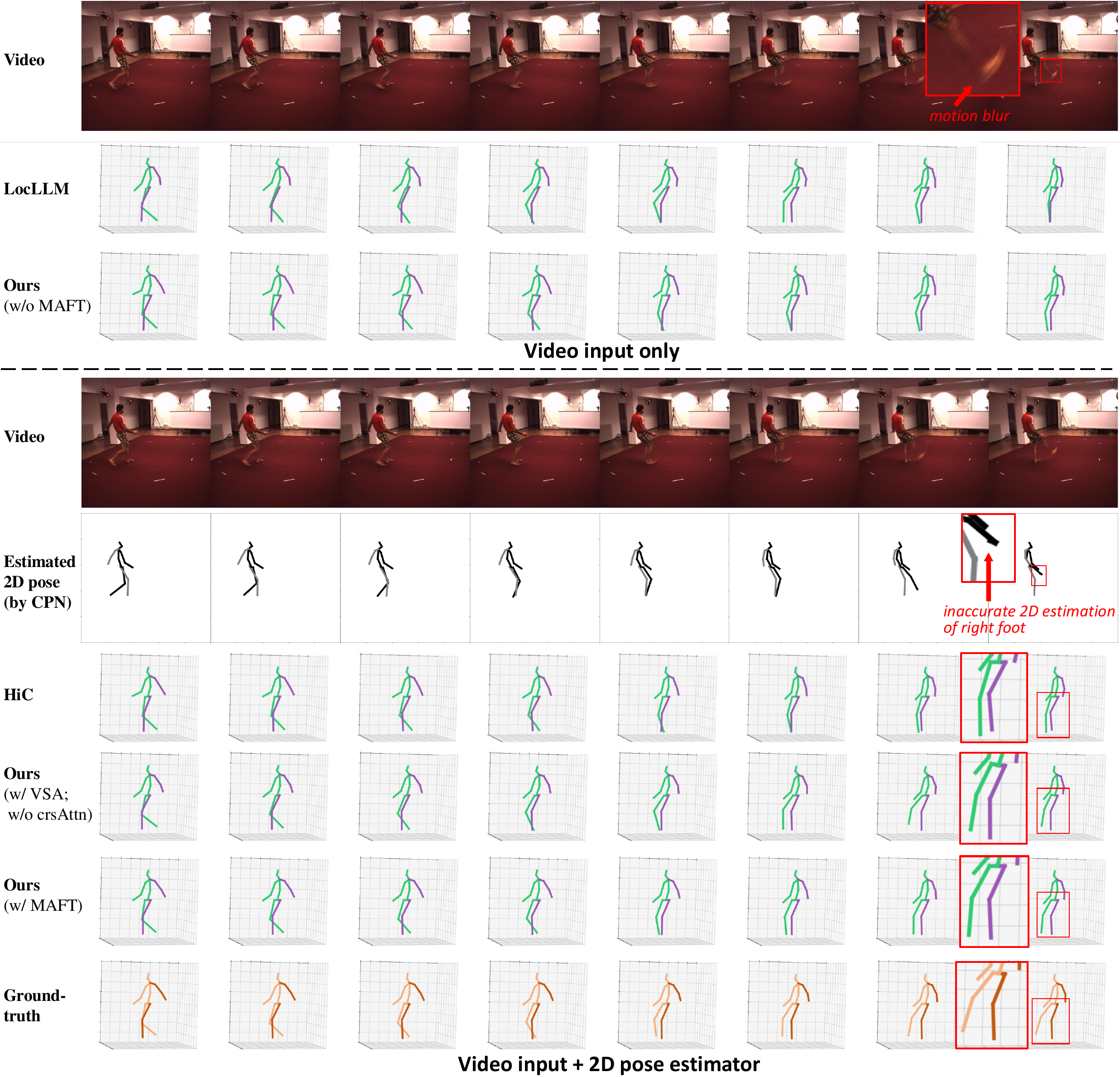}
    \caption{\textbf{Qualitative Results for 3D Pose Estimation on Human3.6M: The Impact of MAFT.} 
    \textbf{Top Block (Video Input Only):} Compared to LocLLM~\cite{wang2024locllm}, our method (even without MAFT) generates more physically plausible poses from raw video, handling motion blur more effectively.
    \textbf{Bottom Block (Video + 2D Pose Input):} This case highlights the robustness of our MAFT module. The off-the-shelf 2D pose estimator (CPN~\cite{chen2018cascaded}) fails to detect the right foot due to occlusion/blur (see red box). Consequently, the baseline HiC~\cite{liu2025human}, which relies heavily on 2D inputs, produces an erroneous pose. In contrast, \textbf{Ours (w/ MAFT)} successfully corrects this error by attending to the visual features via the cross-attention mechanism, recovering the correct leg position that matches the Ground Truth.}
    \label{fig:supp:viz fig_h36m_pe}
   \vspace{-2mm}
\end{figure*}
\begin{figure*}[t]
    \centering
    \includegraphics[width=0.9\textwidth]{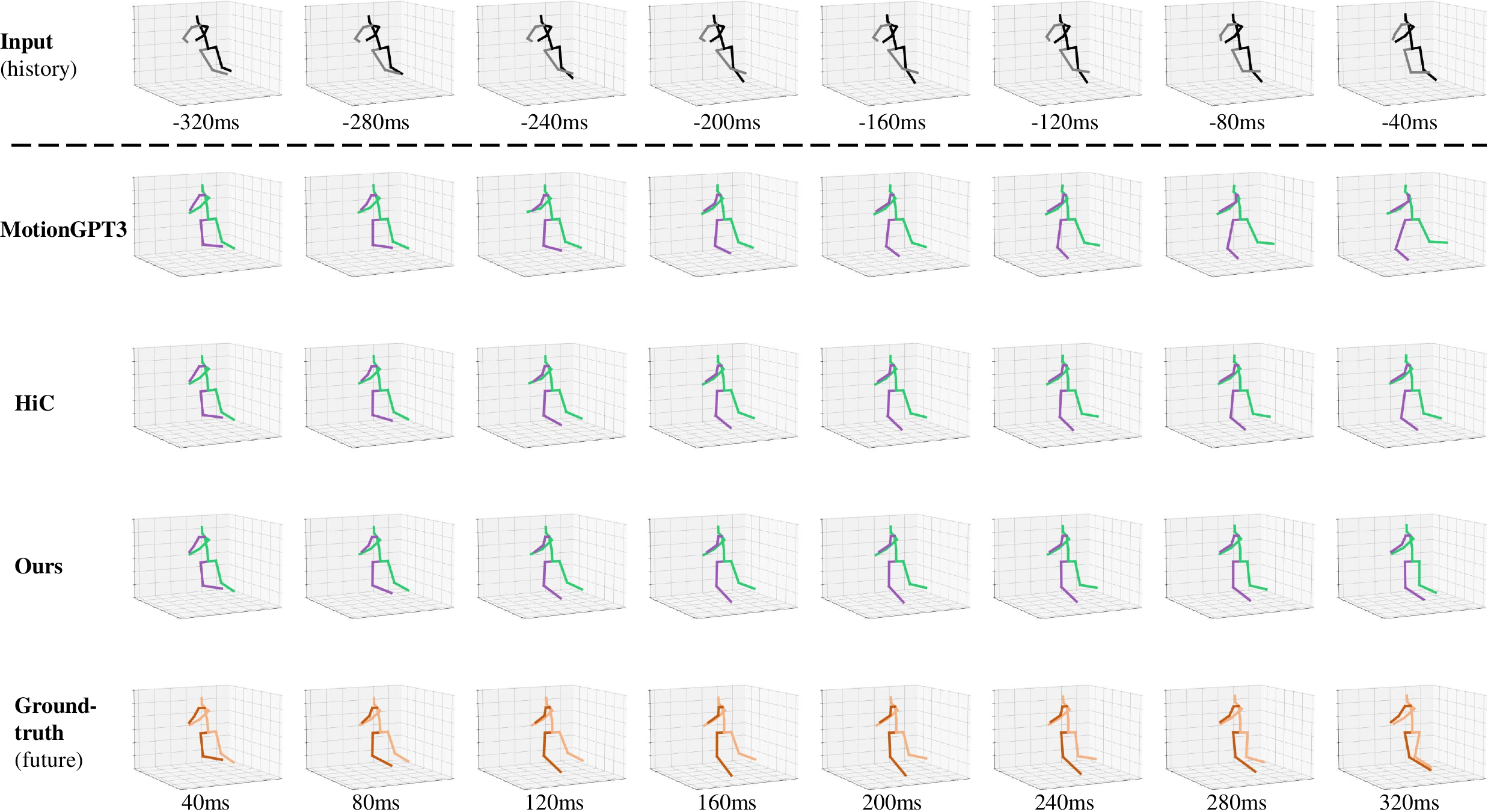}
    \caption{\textbf{Qualitative Comparison for Motion Prediction on Human3.6M.} 
    The model predicts the future 320ms motion given a history sequence (grey). 
    While baselines like MotionGPT3 [36] and HiC [17] exhibit slight temporal jitter or drift in the leg positioning during the "Sitting" action, \textbf{Superman (Ours)} generates a smooth and accurate trajectory that closely aligns with the Ground Truth (orange), demonstrating superior temporal coherence.}
    \label{fig:supp:viz fig_h36m_mp}
   \vspace{-2mm}
\end{figure*}
\begin{figure*}[t]
    \centering
    \includegraphics[width=0.99\textwidth]{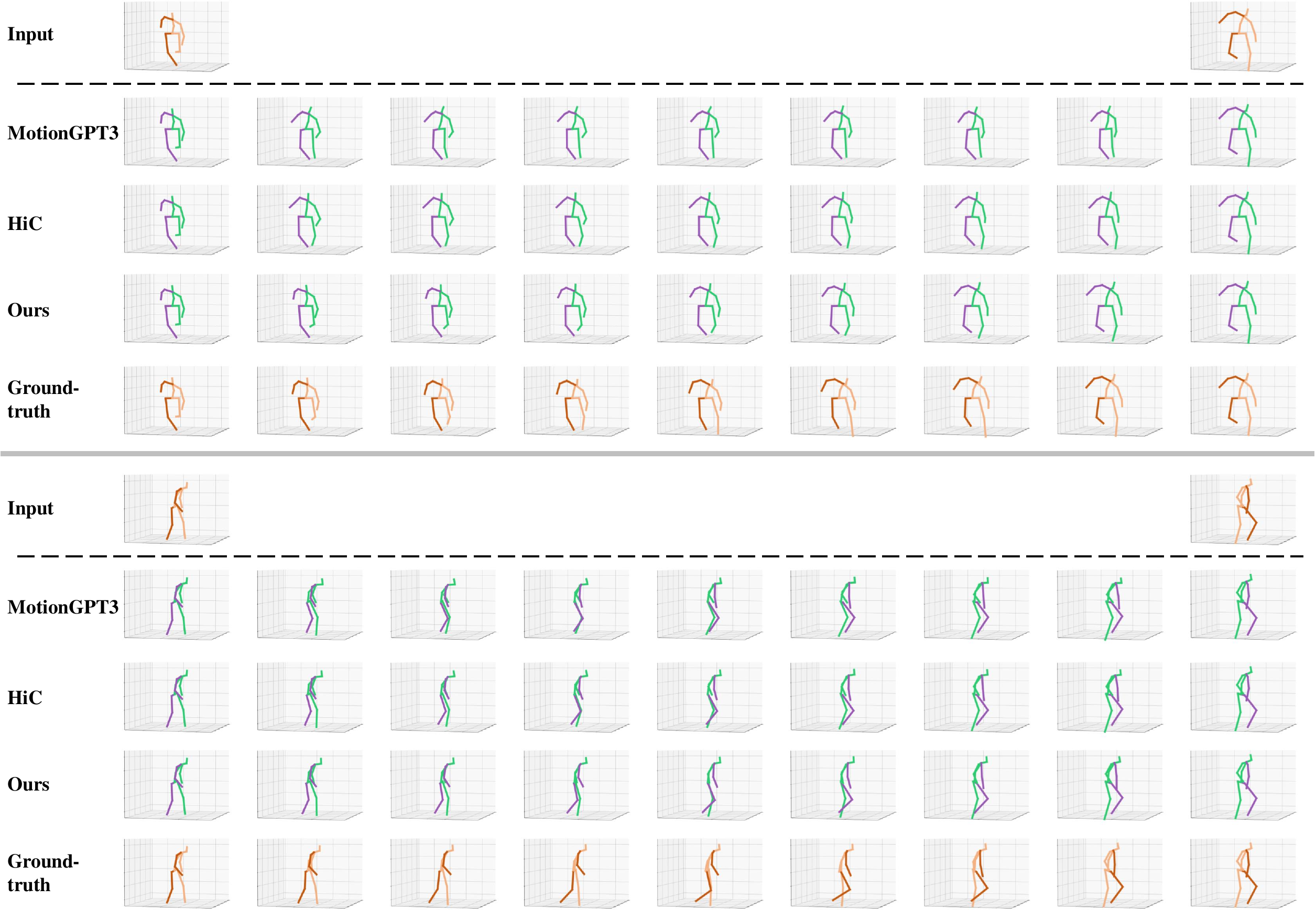}
    \caption{\textbf{Qualitative Results for Motion In-betweening on Human3.6M and 3DPW.} 
    The task is to generate the intermediate motion sequence given only the first and last frames (Input).
    \textbf{Top Row (Human3.6M):} Our model generates a natural transition for the turning action, whereas MotionGPT3 struggles with the limb orientation.
    \textbf{Bottom Row (3DPW):} On the unseen 3DPW dataset, our model demonstrates strong generalization, synthesizing a realistic stepping motion that bridges the gap smoothly, outperforming comparison methods.}
    \label{fig:supp:viz fig_h36m_3dpw_mib}
   \vspace{-2mm}
\end{figure*}

    \subsection{Training Strategy}   
        
        \noindent\textbf{Tokenizer Training:} The VGMT is trained for 500000 steps on Human3.6M. We use the AdamW optimizer with a learning rate of $0.0002$ and batch size $256$.
        
        \noindent\textbf{Unified MLLM Tuning:}
        As shown in Fig.~\ref{fig:supp:viz fig_ft_stage}, unified MLLM tuning includes:
        \textit{1) Multi-Task Instruction Tuning.} We adopt a joint training strategy where data from all three tasks—3D Pose Estimation, Motion Prediction, and Motion In-betweening—are mixed within each batch. To unify these disparate tasks, we utilize the specific instruction templates to convert raw multi-modal data into a standardized "User-Assistant" conversation format. The model is optimized using a standard autoregressive cross-entropy loss over output tokens.
        \textit{2) Optional MAFT Integration.} For tasks involving video input (i.e., 3D Pose Estimation), we incorporate the Motion-Aware Fine-Tuning (MAFT) module. This light-weight component (adding $<0.2\%$ trainable parameters) injects skeletal geometric priors into the visual stream via the Visual-Skeleton Attention (VSA) mechanism, enabling the MLLM to better attend to motion-relevant visual features.
        \textit{3) Hyper-parameters.} The MLLM is fine-tuned using a learning rate of $1e-4$ with a cosine decay scheduler and a warmup ratio of $0.1$. The global batch size is set to $16$. All experiments are conducted on $2$ NVIDIA H20 GPUs using PyTorch.

    \subsection{Instruction Templates}
        To unify diverse motion analysis tasks within a single MLLM framework, we employ instruction tuning. We convert raw data (videos and skeleton sequences) into a standardized conversation format, where the user provides an instruction containing multimodal inputs (e.g., \texttt{<video>}, \texttt{<skeleton>}), and the model generates the corresponding target skeleton tokens.
        Tab.~\ref{tab:supp_prompts} details the specific instruction templates used for each task during the training. 
        \begin{itemize}
            \item \textbf{Vid2Skel (3D Pose Estimation):} The model inputs visual tokens (and optional 2D estimated poses) and is instructed to describe the motion using skeleton tokens.
            \item \textbf{SkelPred (Motion Prediction):} The model inputs a history sequence of skeleton tokens and predicts future frames. Both sequences contain 16 frames (320ms).
            \item \textbf{MotionInBetween (In-betweening):} The model inputs sparse keyframes (start and end) and generates the intermediate transition sequence.
        \end{itemize}

\begin{figure*}[t]
    \centering
    \includegraphics[width=0.85\textwidth]{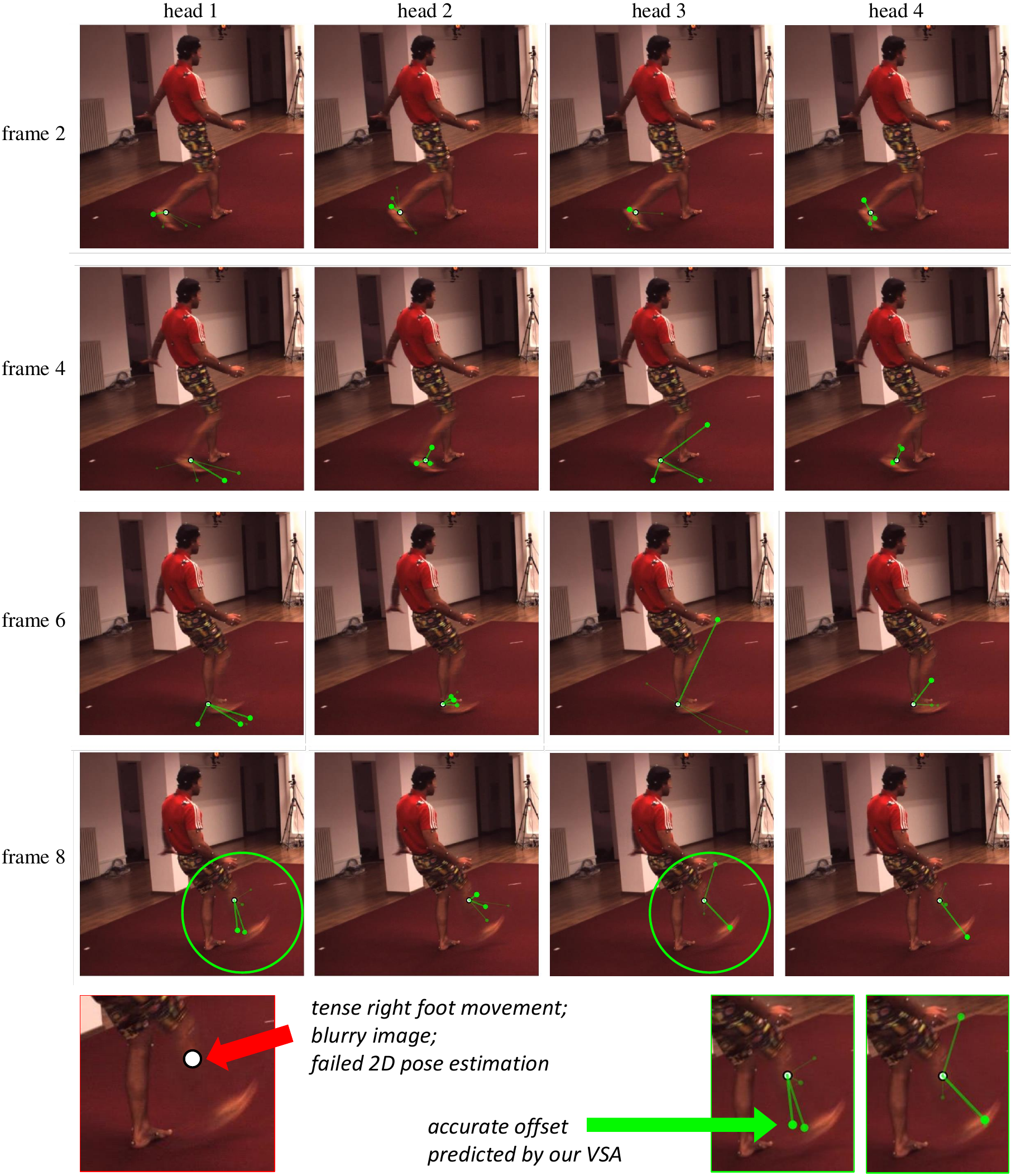}
    \caption{\textbf{Visualization of Adaptive Sampling in Visual-Skeleton Attention (VSA)} -- Right foot as an example. 
    We visualize the learned sampling offsets and weights for different attention heads (columns) across video frames (rows).
    The \textbf{start point} of each green line indicates the initial, potentially noisy 2D keypoint estimated by CPN. 
    The \textbf{green line} represents the predicted offset, pointing to the \textbf{green circle} which denotes the final sampling location. The \textbf{area} of the green circle reflects the attention weight assigned to that sampling point.
    \textbf{Highlight (Bottom Row):} In Frame 8, rapid motion causes the right foot to blur, leading to a failed 2D estimation (white dot). Our VSA module successfully identifies the error, predicting large offsets that redirect attention from the erroneous leg position back to the true, blurry foot region (red and green boxes), thereby ensuring robust feature extraction.}
    \label{fig:supp:viz fig_vsa_v1_s024f000_Rfoot}
   \vspace{-2mm}
\end{figure*}
\begin{figure*}[t]
    \centering
    \includegraphics[width=0.7\textwidth]{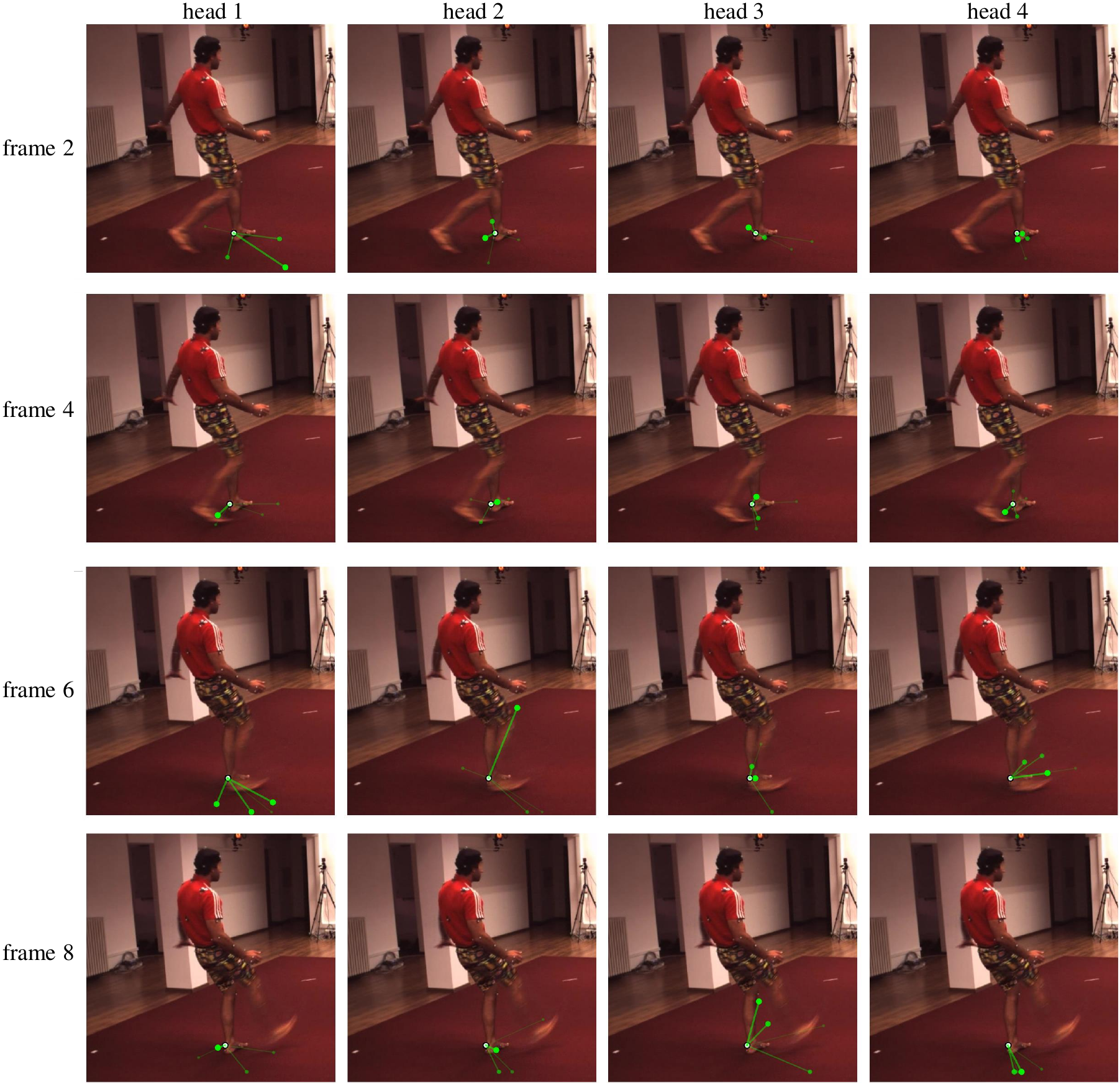}
    \caption{\textbf{Visualization of Adaptive Sampling in Visual-Skeleton Attention (VSA)} -- Left foot as an example.
    The VSA sampling process for the planting and lifting phase of the left foot is illustrated.
    \textbf{Multi-Head Diversity:} Different attention heads exhibit diverse sampling strategies. For instance, while Head 2 tends to focus tightly on the joint center (small offsets), Head 4 explores a wider neighborhood (larger offsets) to capture contextual motion cues. This diversity allows the model to build a robust representation of the limb's state even during complex articulation.}
    \label{fig:supp:viz fig_vsa_v1_s024f000_Lfoot}
   \vspace{-2mm}
\end{figure*}

\begin{figure*}[t]
    \centering
    \includegraphics[width=0.95\textwidth]{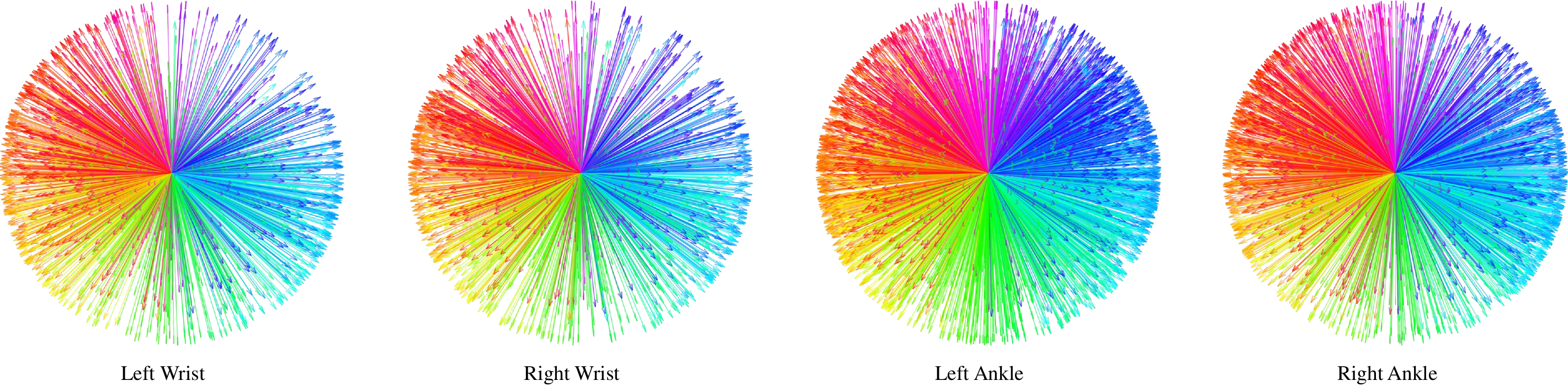}
    \caption{\textbf{Visualization of the ``Semantic Spheres'' of Motion Primitives across Different Joints.} 
    Each sphere visualizes the discrete vocabulary learned by our Vision-Guided Motion Tokenizer (VGMT) for a specific body joint. 
    Each fiber-like line represents a unique code in the codebook, plotted as its normalized average displacement vector (starting from the origin). The color indicates the motion direction (azimuthal hue).
    \textbf{(1) Omnidirectional Coverage:} The dense, isotropic spherical distribution demonstrates that the codebook effectively covers the full manifold of possible motion directions without mode collapse.
    \textbf{(2) High-Fidelity Granularity:} The high density of vectors (e.g., over 2,000 active codes for hands and feet) confirms that the tokenizer captures fine-grained kinematic details.
    }
    \label{fig:supp:viz fig_vqvae_semantics}
   \vspace{-2mm}
\end{figure*}

\section{Additional Visualization Results}

        We further demonstrate the superiority of Superman through qualitative comparisons across tasks.
        
        \subsection{Robustness in 3D Pose Estimation}
        Fig.~\ref{fig:supp:viz fig_h36m_pe} presents a visual comparison for pose estimation.
        \begin{itemize}
            \item \textbf{Correction of Upstream Errors:} The bottom block of Fig.~\ref{fig:supp:viz fig_h36m_pe} highlights a challenging scenario where the off-the-shelf 2D pose estimator (CPN~\cite{chen2018cascaded}) fails to detect the right foot due to motion blur (see red box). Baseline methods like HiC~\cite{liu2025human}, which rely heavily on 2D keypoints, propagate this error into the final 3D output.
            \item \textbf{Effect of MAFT:} Our model with Motion-Aware Fine-Tuning (MAFT) effectively mitigates this issue. By leveraging the Cross-Attention mechanism to query visual features directly from the video, \textit{Superman} ``looks back'' at the raw frames to recover the missing foot information.
        \end{itemize}
        
        \subsection{Temporal Coherence in Motion Prediction}
        Fig.~\ref{fig:supp:viz fig_h36m_mp} compares the prediction capabilities on the Human3.6M dataset. The task involves predicting 320ms of future motion based on historical context. 
        As observed in the "Sitting" sequence, baseline methods such as MotionGPT3~\cite{zhu2025motiongpt3} and HiC~\cite{liu2025human} tend to generate stiff or slightly drifting motions as time progresses. In contrast, \textit{Superman} maintains high temporal fidelity, accurately predicting the descent of the body and the knee bending trajectory, closely matching the ground truth.
        
        \subsection{Generalization in Motion In-betweening}
        Fig.~\ref{fig:supp:viz fig_h36m_3dpw_mib} evaluates the Motion In-betweening task, where the model must synthesize a coherent sequence connecting a start and an end frame.
        \begin{itemize}
            \item \textbf{In-Domain (Human3.6M):} In the top row, our model generates a smooth turning motion, naturally interpolating the limb rotations.
            \item \textbf{Out-of-Domain (3DPW):} The bottom row demonstrates generalization on the unseen 3DPW dataset. Despite never being trained on this data, \textit{Superman} synthesizes a realistic stepping motion. Comparison methods often struggle with such large gaps or unseen poses, resulting in artifacts. Our success here attributes to the robust, vision-grounded motion vocabulary learned by the VGMT.
        \end{itemize}

\section{Visualization of Intermediate Features}
\label{sec:supp_chap3_intermediate}

    We analyze \textit{Superman} by visualizing the intermediate representations learned by its key components. This provides qualitative evidence for the mechanisms of VSA sampling, MAFT attention, and the semantic structure of the learned codebook.

    \subsection{Adaptive Sampling in VSA}
        A critical challenge in vision-guided motion generation is the unreliability of upstream 2D pose estimators, particularly under rapid motion or occlusion. To validate how our Visual-Skeleton Attention (VSA) module addresses this, we visualize the internal sampling process in Fig.~\ref{fig:supp:viz fig_vsa_v1_s024f000_Rfoot} \& \ref{fig:supp:viz fig_vsa_v1_s024f000_Lfoot}.
        
        \textbf{Visualization Protocol:} 
        We visualize the sampling behavior of different attention heads across temporal frames. In the visualization:
        \begin{itemize}
            \item The \textbf{Start Point} of a green line represents the initial reference point provided by the off-the-shelf 2D pose estimator (CPN).
            \item The \textbf{Green Line} represents the learned offset vector ($\Delta p$) predicted by the VSA module.
            \item The \textbf{End Point (Green Circle)} indicates the final adaptive sampling location on the feature map.
            \item The \textbf{Area of the Circle} corresponds to the learned attention weight, where larger circles indicate higher contribution to the aggregated feature.
        \end{itemize}
        
        \textbf{Analysis of Correction Capability:}
        As highlighted in the bottom row of Fig.~\ref{fig:supp:viz fig_vsa_v1_s024f000_Rfoot} (Frame 8), the fast-moving right foot causes significant motion blur, leading the fixed 2D estimator to incorrectly localize the joint on the leg or background (see the white dot). 
        However, our VSA module detects this semantic misalignment. The predicted offsets (green lines) diverge from the erroneous initial point and accurately point towards the actual "ghosting" region of the blurry foot. Furthermore, different attention heads (Head 1-4) learn to focus on complementary features—some attending to the heel and others to the toe trajectory—aggregating a robust visual representation despite the noisy input. This confirms that VSA functions not just as a sampler, but as a dynamic \textit{visual corrector}.
        
        Fig.~\ref{fig:supp:viz fig_vsa_v1_s024f000_Lfoot} demonstrates the tracking of the left foot. Even when the 2D estimation is relatively stable, the VSA module actively refines the sampling points. Different attention heads (Head 1-4) can be seen focusing on distinct semantic parts (e.g., heel vs. toe) or expanding the receptive field to capture context, ensuring a rich feature representation.

    \subsection{Semantic Analysis of the Codebook}
        Finally, to validate that our discrete tokens carry explicit physical meanings, we perform a "Semantic Sphere" analysis. For every code index $k$ in the learned vocabulary, we calculate the average displacement vector of all motion segments assigned to it. We then visualize these vectors as fiber-like lines radiating from the origin, normalized to unit length to emphasize directionality.
        
        Fig.~\ref{fig:supp:viz fig_vqvae_semantics} presents the results for four critical end-effectors: \textbf{Left/Right Wrists} and \textbf{Left/Right Ankles}. These joints typically exhibit the highest degrees of freedom and complexity in human motion.
        Key observations:
        \begin{itemize}
            \item \textbf{Isotropic Coverage (No Mode Collapse):} As illustrated in the figure, the motion vectors for all four joints form dense, almost perfect spheres. The uniform distribution of vectors in every direction (isotropic) provides compelling evidence that our codebook effectively covers the full manifold of possible motion directions. There are no significant "holes" or gaps, indicating that the model has avoided mode collapse and can generate diverse motions in any 3D direction.
            
            \item \textbf{High-Fidelity Granularity:} The sheer density of the fibers—representing over 2,000 active codes for these end-effectors—confirms that the tokenizer captures fine-grained kinematic details. The model allocates a vast vocabulary to describe subtle variations in hand and foot movements, which is crucial for high-quality motion generation.
            
            \item \textbf{Directional Semantics:} The smooth transition of colors (representing azimuthal direction) confirms that the codes are semantically organized. Specific codes map uniquely to specific physical directions, allowing the subsequent MLLM to control motion with precise directional intent.
        \end{itemize}

%% file: main.bib
@String(IJCV = {Int. J. Comput. Vis.})

@String(CVPR= {IEEE Conf. Comput. Vis. Pattern Recog.})

@String(ICCV= {Int. Conf. Comput. Vis.})

@String(ECCV= {Eur. Conf. Comput. Vis.})

@String(TOG= {ACM Trans. Graph.})

@String(AAAI = {AAAI})

@String(IJCV  = {IJCV})

@String(CVPR  = {CVPR})

@String(ICCV  = {ICCV})

@String(ECCV  = {ECCV})

@String(TOG   = {ACM TOG})

@article{ding2025mtvcrafter,
  title={MTVCrafter: 4D Motion Tokenization for Open-World Human Image Animation},
  author={Ding, Yanbo and Hu, Xirui and Guo, Zhizhi and Zhang, Chi and Wang, Yali},
  journal={arXiv preprint arXiv:2505.10238},
  year={2025}
}

@article{liu2024point,
  title={Point-in-context: Understanding point cloud via in-context learning},
  author={Liu, Mengyuan and Fang, Zhongbin and Li, Xia and Buhmann, Joachim M and Ye, Deheng and Li, Xiangtai and Loy, Chen Change},
  journal={IJCV},
  year={2026}
}

@inproceedings{chen2018cascaded,
  title={Cascaded pyramid network for multi-person pose estimation},
  author={Chen, Yilun and Wang, Zhicheng and Peng, Yuxiang and Zhang, Zhiqiang and Yu, Gang and Sun, Jian},
  booktitle={CVPR},
  pages={7103--7112},
  year={2018}
}

@inproceedings{bogo2016keep,
  title={Keep it SMPL: automatic estimation of 3D human pose and shape from a single image},
  author={Bogo, Federica and Kanazawa, Angjoo and Lassner, Christoph and Gehler, Peter and Romero, Javier and Black, Michael J},
  booktitle={ECCV},
  year={2016},
}

@article{zhu2025motiongpt3,
  title={MotionGPT3: Human Motion as a Second Modality},
  author={Zhu, Bingfan and Jiang, Biao and Wang, Sunyi and Tang, Shixiang and Chen, Tao and Luo, Linjie and Zheng, Youyi and Chen, Xin},
  journal={arXiv preprint arXiv:2506.24086},
  year={2025}
}

@inproceedings{wang2024locllm,
    title     = {LocLLM: Exploiting Generalizable Human Keypoint Localization via Large Language Model},
    author    = {Wang, Dongkai and Xuan, Shiyu and Zhang, Shiliang},
    booktitle = {CVPR},
    year      = {2024}
}

@inproceedings{von2018_3dpw,
  title={Recovering accurate 3d human pose in the wild using imus and a moving camera},
  author={Von Marcard, Timo and Henschel, Roberto and Black, Michael J and Rosenhahn, Bodo and Pons-Moll, Gerard},
  booktitle={ECCV},
  year={2018}
}

@article{fang2023explore,
  title={Explore in-context learning for 3d point cloud understanding},
  author={Fang, Zhongbin and Li, Xiangtai and Li, Xia and Buhmann, Joachim M and Loy, Chen Change and Liu, Mengyuan},
  booktitle={NeurIPS},
  year={2023}
}

@article{liu2025human,
  title={Human-in-Context: Unified Cross-Domain 3D Human Motion Modeling via In-Context Learning},
  author={Liu, Mengyuan and Wang, Xinshun and Fang, Zhongbin and Ye, Deheng and Li, Xia and Tang, Tao and Wu, Songtao and Li, Xiangtai and Yang, Ming-Hsuan},
  journal={arXiv preprint arXiv:2508.10897},
  year={2025}
}

@article{ionescu2013h36m,
  title={Human3. 6m: large scale datasets and predictive methods for 3d human sensing in natural environments},
  author={Ionescu, Catalin and Papava, Dragos and Olaru, Vlad and Sminchisescu, Cristian},
  journal={IEEE T-PAMI},
  year={2013},
}

@inproceedings{wang2024sic,
  title={Skeleton-in-context: unified skeleton sequence modeling with in-context learning},
  author={Wang, Xinshun and Fang, Zhongbin and Li, Xia and Li, Xiangtai and Chen, Chen and Liu, Mengyuan},
  booktitle={CVPR},
  year={2024}
}

@inproceedings{zhu2023motionbert,
  title={MotionBERT: a Unified Perspective on Learning Human Motion Representations},
  author={Zhu, Wentao and Ma, Xiaoxuan and Liu, Zhaoyang and Liu, Libin and Wu, Wayne and Wang, Yizhou},
  booktitle={ICCV},
  year={2023}
}

@inproceedings{feng2025posellava,
  title={PoseLLaVA: Pose Centric Multimodal LLM for Fine-Grained 3D Pose Manipulation},
  author={Feng, Dong and Guo, Ping and Peng, Encheng and Zhu, Mingmin and Yu, Wenhao and Wang, Peng},
  booktitle={AAAI},
  year={2025}
}

@article{zhang2025llava,
  title={LLaVA-Pose: Enhancing Human Pose and Action Understanding via Keypoint-Integrated Instruction Tuning},
  author={Zhang, Dewen and Hussain, Tahir and An, Wangpeng and Shouno, Hayaru},
  journal={arXiv preprint arXiv:2506.21317},
  year={2025}
}

@article{chen2024motionllm,
  title={Motionllm: Understanding human behaviors from human motions and videos},
  author={Chen, Ling-Hao and Lu, Shunlin and Zeng, Ailing and Zhang, Hao and Wang, Benyou and Zhang, Ruimao and Zhang, Lei},
  journal={arXiv preprint arXiv:2405.20340},
  year={2024}
}

@article{jiang2023motiongpt,
  title={Motiongpt: Human motion as a foreign language},
  author={Jiang, Biao and Chen, Xin and Liu, Wen and Yu, Jingyi and Yu, Gang and Chen, Tao},
  booktitle={NeurIPS},
  year={2023}
}

@inproceedings{li2025unipose,
  title={Unipose: A unified multimodal framework for human pose comprehension, generation and editing},
  author={Li, Yiheng and Hou, Ruibing and Chang, Hong and Shan, Shiguang and Chen, Xilin},
  booktitle={CVPR},
  year={2025}
}

@inproceedings{feng2024chatpose,
  title={Chatpose: Chatting about 3d human pose},
  author={Feng, Yao and Lin, Jing and Dwivedi, Sai Kumar and Sun, Yu and Patel, Priyanka and Black, Michael J},
  booktitle={CVPR},
  year={2024}
}

@inproceedings{sinha2025ski,
  title={SKI Models: Skeleton Induced Vision-Language Embeddings for Understanding Activities of Daily Living},
  author={Sinha, Arkaprava and Reilly, Dominick and Bremond, Francois and Wang, Pu and Das, Srijan},
  booktitle={AAAI},
  year={2025}
}

@inproceedings{andriluka2018posetrack,
  title={Posetrack: a benchmark for human pose estimation and tracking},
  author={Andriluka, Mykhaylo and Iqbal, Umar and Insafutdinov, Eldar and Pishchulin, Leonid and Milan, Anton and Gall, Juergen and Schiele, Bernt},
  booktitle={CVPR},
  year={2018}
}

@inproceedings{zhang2020distribution,
  title={Distribution-aware coordinate representation for human pose estimation},
  author={Zhang, Feng and Zhu, Xiatian and Dai, Hanbin and Ye, Mao and Zhu, Ce},
  booktitle={CVPR},
  year={2020}
}

@inproceedings{liu2018pem,
  title={Recognizing Human Actions as the Evolution of Pose Estimation Maps},
  author={Liu, Mengyuan and Yuan, Junsong},
  booktitle={CVPR},
  year={2018}
}

@article{zhang2024pose,
  title={Pose Magic: efficient and Temporally Consistent Human Pose Estimation with a Hybrid Mamba-GCN Network},
  author={Zhang, Xinyi and Bao, Qiqi and Cui, Qinpeng and Yang, Wenming and Liao, Qingmin},
  journal={arXiv preprint arXiv:2408.02922},
  year={2024}
}

@inproceedings{cui2021towards,
  title={Towards accurate 3d human motion prediction from incomplete observations},
  author={Cui, Qiongjie and Sun, Huaijiang},
  booktitle={CVPR},
  year={2021}
}

@article{wang2024dynamic,
  title={Dynamic dense graph convolutional network for skeleton-based human motion prediction},
  author={Wang, Xinshun and Zhang, Wanying and Wang, Can and Gao, Yuan and Liu, Mengyuan},
  journal={IEEE T-IP}, 
  year={2024}
}

@inproceedings{wang2024gcnext,
  title={GCNext: towards the Unity of graph convolutions for human motion prediction},
  author={Wang, Xinshun and Cui, Qiongjie and Chen, Chen and Liu, Mengyuan},
  booktitle={AAAI},
  year={2024}
}

@article{zhou2020generative,
  title={Generative tweening: long-term inbetweening of 3d human motions},
  author={Zhou, Yi and Lu, Jingwan and Barnes, Connelly and Yang, Jimei and Xiang, Sitao and others},
  journal={arXiv preprint arXiv:2005.08891},
  year={2020}
}

@inproceedings{kaufmann2020convolutional,
  title={Convolutional autoencoders for human motion infilling},
  author={Kaufmann, Manuel and Aksan, Emre and Song, Jie and Pece, Fabrizio and Ziegler, Remo and Hilliges, Otmar},
  booktitle={3DV},
  year={2020},
}

@inproceedings{hernandez2019human,
  title={Human motion prediction via spatio-temporal inpainting},
  author={Hernandez, Alejandro and Gall, Jurgen and Moreno-Noguer, Francesc},
  booktitle={ICCV},
  year={2019}
}

@article{harvey2020robust,
  title={Robust motion in-betweening},
  author={Harvey, F{\'e}lix G and Yurick, Mike and Nowrouzezahrai, Derek and Pal, Christopher},
  journal={ACM TOG},
  year={2020},
}

@article{toshev2014deeppose,
  title={DeepPose: Human Pose Estimation via Deep Neural Networks},
  author={Alexander Toshev and Christian Szegedy},
  booktitle={CVPR},
  year={2013},
}

@article{martinez2017simple,
  title={A Simple Yet Effective Baseline for 3d Human Pose Estimation},
  author={Julieta Martinez and Rayat Hossain and Javier Romero and J. Little},
  booktitle={ICCV},
  year={2017},
}

@article{zheng2021poseformer,
  title={3D Human Pose Estimation with Spatial and Temporal Transformers},
  author={Ce Zheng and Sijie Zhu and Mat'ias Mendieta and Taojiannan Yang and Chen Chen and Zhengming Ding},
  booktitle={ICCV},
  year={2021},
}

@article{pavllo20193d,
  title={3D Human Pose Estimation in Video With Temporal Convolutions and Semi-Supervised Training},
  author={Dario Pavllo and Christoph Feichtenhofer and David Grangier and Michael Auli},
  booktitle={CVPR},
  year={2018},
}

@article{van2017neural,
  title={Neural discrete representation learning},
  author={Van Den Oord, Aaron and Vinyals, Oriol and others},
  booktitle={NeurIPS},
  volume={30},
  year={2017}
}

@article{liu2023visual,
  title={Visual instruction tuning},
  author={Liu, Haotian and Li, Chunyuan and Wu, Qingyang and Lee, Yong Jae},
  booktitle={NeurIPS},
  volume={36},
  pages={34892--34916},
  year={2023}
}

@article{ye2023mplug,
  title={mplug-owl: Modularization empowers large language models with multimodality},
  author={Ye, Qinghao and Xu, Haiyang and Xu, Guohai and Ye, Jiabo and Yan, Ming and Zhou, Yiyang and Wang, Junyang and Hu, Anwen and Shi, Pengcheng and Shi, Yaya and others},
  journal={arXiv preprint arXiv:2304.14178},
  year={2023}
}

@article{dai2023instructblip,
  title={Instructblip: Towards general-purpose vision-language models with instruction tuning},
  author={Dai, Wenliang and Li, Junnan and Li, Dongxu and Tiong, Anthony and Zhao, Junqi and Wang, Weisheng and Li, Boyang and Fung, Pascale N and Hoi, Steven},
  booktitle={NeurIPS},
  year={2023}
}

@article{bai2025qwen2,
  title={Qwen2. 5-vl technical report},
  author={Bai, Shuai and Chen, Keqin and Liu, Xuejing and Wang, Jialin and Ge, Wenbin and Song, Sibo and Dang, Kai and Wang, Peng and Wang, Shijie and Tang, Jun and others},
  journal={arXiv preprint arXiv:2502.13923},
  year={2025}
}

@inproceedings{sun2019deep,
  title={Deep high-resolution representation learning for human pose estimation},
  author={Sun, Ke and Xiao, Bin and Liu, Dong and Wang, Jingdong},
  booktitle={CVPR},
  year={2019}
}

@article{zhu2020deformable,
  title={Deformable detr: Deformable transformers for end-to-end object detection},
  author={Zhu, Xizhou and Su, Weijie and Lu, Lewei and Li, Bin and Wang, Xiaogang and Dai, Jifeng},
  journal={arXiv preprint arXiv:2010.04159},
  year={2020}
}
